\documentclass[sigconf]{acmart}

\AtBeginDocument{%
  }

\usepackage{booktabs}
\usepackage{multirow}
\usepackage{graphicx}
\usepackage{xcolor}
\usepackage{enumitem}
\usepackage{algorithm}
\usepackage{algorithmic}

\copyrightyear{2026}
\acmYear{2026}
\setcopyright{cc}
\setcctype{by}
\acmConference[KDD '26]{Proceedings of the 32nd ACM SIGKDD Conference on Knowledge Discovery and Data Mining V.2}{August 09--13, 2026}{Jeju Island, Republic of Korea}
\acmBooktitle{Proceedings of the 32nd ACM SIGKDD Conference on Knowledge Discovery and Data Mining V.2 (KDD '26), August 09--13, 2026, Jeju Island, Republic of Korea}
\acmDOI{10.1145/3770855.3818022}
\acmISBN{979-8-4007-2259-2/2026/08}

\begin{document}

\title{Frequency-Domain Multi-Modality Transportation Modeling}

\author{Jiewen Deng}
  \orcid{0000-0002-6172-4390}
  \email{dengjw1@outlook.com}
  \affiliation{%
    \institution{Southern University of \\Science and Technology}
    \city{Shenzhen}
    \country{China}
  }

  \author{Hangchen Liu}
  \orcid{0009-0000-9220-522X}
  \email{liuhc3@outlook.com}
  \affiliation{
    \institution{The University of Tokyo}
    \city{Tokyo}
    \country{Japan}
  }

  \author{Junchen Li}
  \orcid{0009-0007-5222-7679}
  \email{bobjunchen2@gmail.com}
  \affiliation{%
    \institution{Southern University of \\Science and Technology}
    \city{Shenzhen}
    \country{China}
  }

  \author{Boyuan Zhang}
  \orcid{0009-0001-2755-7525}
  \email{zby9973@outlook.com}
  \affiliation{
    \institution{The University of Tokyo}
    \city{Tokyo}
    \country{Japan}
  }

  \author{Renhe Jiang}
  \authornote{Corresponding author.}
  \orcid{0000-0003-2593-4638}
  \email{jiangrh@csis.u-tokyo.ac.jp}
  \affiliation{%
    \institution{The University of Tokyo}
    \city{Tokyo}
    \country{Japan}
  }

\begin{abstract}
Multi-modality transportation refers to urban systems composed of multiple transportation modes, such as traffic flow and public transit, whose dynamics are coupled by shared temporal patterns. Accurate multi-modality transportation forecasting remains challenging because (1) different modalities exhibit distinct spectral characteristics and (2) interact unevenly across frequencies, whereas most existing methods operate primarily in the time domain or rely on coarse feature fusion. 
To address these limitations, we propose a lightweight yet effective \textbf{Fre}quency-Domain Multi-\textbf{Mo}dality modeling (\textbf{FreMo}) that explicitly exploits the frequency domain to enable adaptive and selective cross-modality synergy. FreMo disentangles modality-wise spectral refinement from cross-modality synergy and supports plug-and-play integration with general time series backbones. Specifically, FreMo introduces a \emph{Modality-Wise Frequency Filter} (MFF) to adaptively refine spectral components within each modality, emphasizing informative frequencies while suppressing noise. FreMo further incorporates a \emph{Frequency-Guided Synergy Integrator} (FSI) that selectively aggregates information across modalities based on their relative contribution at each frequency, facilitating effective cross-modality knowledge sharing while mitigating negative transfer. Extensive experiments on real-world datasets show that FreMo consistently outperforms state-of-the-art baselines, with superior performance and generalization across diverse forecasting scenarios.
The code is available at \url{https://github.com/beginner-sketch/FreMo}.
\end{abstract}

\ccsdesc[500]{Information systems~Spatial-temporal systems}
\ccsdesc[500]{Computing methodologies~Artificial intelligence}

\keywords{frequency-domain, multi-modality transportation modeling}

\maketitle

\begin{figure}[h]
  \centering
  \includegraphics[width=\linewidth]{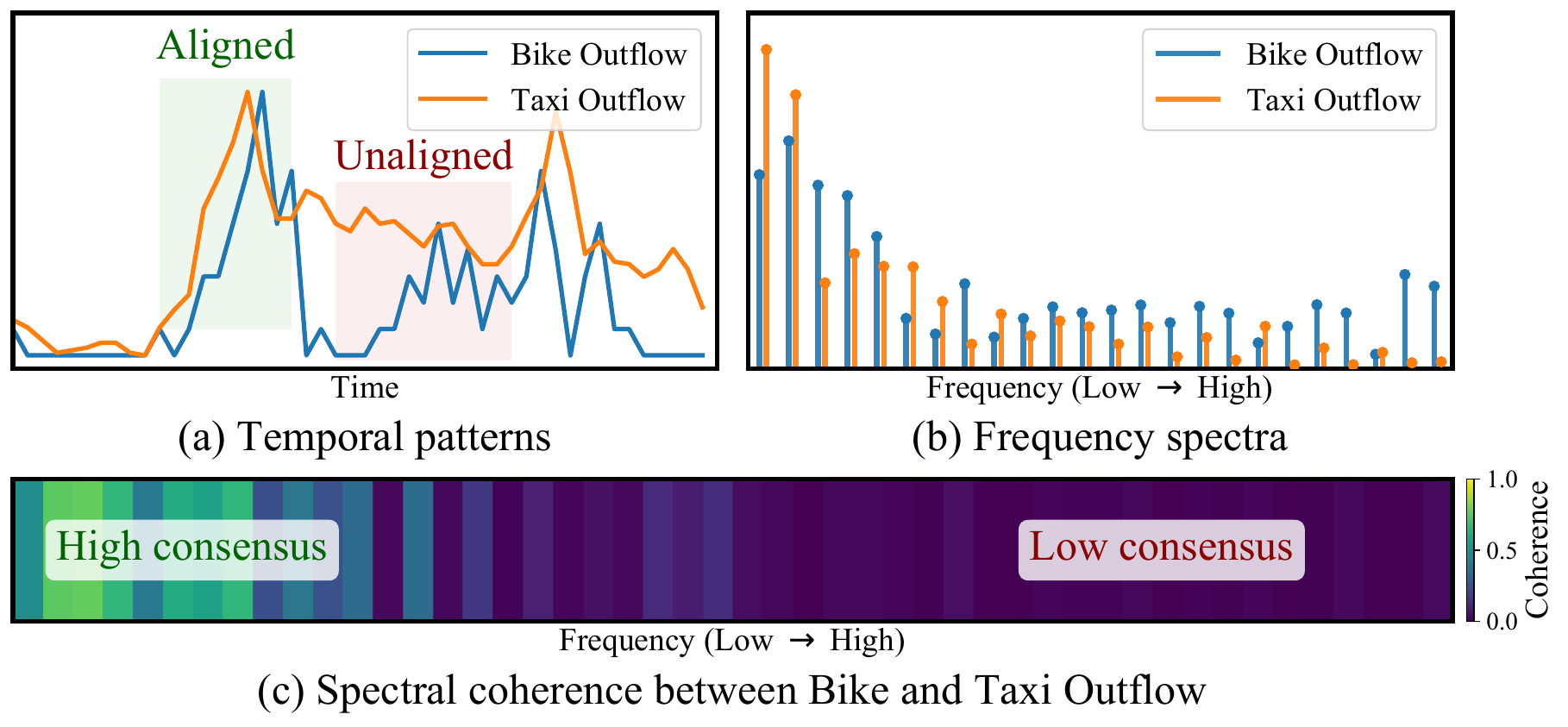}
  \caption{Multi-modality transportation data with representative modalities (Bike and Taxi Outflow). (a) Temporal patterns exhibit aligned trends but mismatched details. (b) Frequency spectra differ across modalities. (c) Spectral coherence is high at low frequencies but drops at high frequencies.}
  \Description{A figure contains the temporal patterns, frequency-domain profiles, and the frequency-wise cross-modality coherence of two different modalities.}
  \label{figure:intro}
\end{figure}

\section{Introduction}
Multi-modality transportation data are structured time series that record traffic dynamics over time across multiple spatial units and transportation modes, such as public transit and bike-sharing systems. At each timestamp, observations from different modalities (e.g., Bike and Taxi flows~\cite{zhang2023autost,zhang2017stresnet,jiang2021deepcrowd}) describe complementary aspects of traffic state. Recently, time series analysis has received increasing attention due to its vital role in urban computing for various downstream tasks \cite{liu2024elephant,wang2023drift,lu2023outofdistribution,cheng2023formertime,huang2024hdmixer,yu2024ginar,chen2024pathformer,dai2024periodicity,lin2024cyclenet,zhao2024lift,qiu2025duet}.
Within this field, multi-modality time series modeling has become increasingly important for its ability to integrate diverse information sources and improve predictive performance. It has been widely applied in real-world scenarios such as transportation forecasting~\cite{wang2022event,deng2024mossl,deng2023tts,zhou2018attconvlstm,deng2023gmrl}, crime analysis~\cite{huang2018deepcrime,xia2021stshn,huang2019mist}, and air quality monitoring~\cite{deng2023tts,deng2023gmrl,deng2024mossl,han2021joint}. 

However, effectively coordinating multiple modalities for reliable forecasting remains challenging.
A central goal of multi-modality learning is to capture shared patterns across modalities while preserving modality-specific characteristics, enabling complementary information exchange without introducing interference. 
Real-world data exhibit complex and diverse temporal dynamics across modalities \cite{deng2023gmrl,deng2024mossl}. Different modalities are informative at different temporal scales: some mainly reflect long-term trends, while others are more sensitive to short-term fluctuations. 
These limitations further expose three key challenges across temporal and frequency dimensions, as illustrated in Figure~\ref{figure:intro}. 

\textbf{First, rigid temporal alignment fails to capture fine-grained cross-modality differences.} In the time domain (Figure~\ref{figure:intro}(a)), both modalities exhibit aligned trends driven by daily commuting patterns (e.g., morning and evening peaks). At finer time resolutions, clear unaligned details emerge. Bike Outflow shows sharper fluctuations and intermittent drops that do not align well with Taxi Outflow. 
Most existing methods~\cite{fang2021mdtp,huang2019mist,wu2020hierarchically,ye2019co} primarily fuse modalities in the time domain and apply a uniform fusion rule across time steps. This scale-agnostic design may enforce coarse cross-modal consistency, but it cannot separate scale-dependent dynamics or identify when each modality should dominate, often leading to misaligned interactions, noisy information exchange, and limited gains.
To explicitly separate temporal scales, a natural way is to move from the time domain to the frequency domain, where different scales can be characterized by distinct frequency components.

\textbf{Second, distinct spectral compositions across modalities challenge uniform modeling.} Figure~\ref{figure:intro}(b) reveals clearer modality differences in the frequency domain.
At low frequencies, modalities exhibit strong spectral energy, indicating consistent daily patterns. At higher frequencies, their spectra diverge with modality-specific peaks and distinct decay behaviors, suggesting that modality contributions vary across frequency bands. 
However, existing frequency-domain methods~\cite{zhou2022fedformer,yue2025freeformer,piao2024fredformer,yi2023frequency,ye2024frequency,fu2025frequency} are modality-agnostic, leaving the modality-frequency relationship (i.e., which spectral components are informative for each modality) underexplored.

\textbf{Third, indiscriminate collaboration obscures reliable cross-modality consensus signals.} Spectral coherence (Figure~\ref{figure:intro}(c)) measures dependency between modalities at different frequencies. \emph{High Consensus} appears in low-frequency regions, indicating reliable shared dynamics. In contrast, coherence drops at high frequencies, marking a \emph{Low Consensus} area dominated by modality-specific variations. This suggests that cross-modality collaboration should be selective and conditioned on frequency-dependent reliability, rather than uniformly enforced across components. We refer to selective and reliability-aware collaboration as \emph{synergy} across modalities.

To address these challenges, we propose a novel \textbf{Fre}quency-Domain Multi-\textbf{Mo}dality modeling framework (\textbf{FreMo}) for transportation forecasting. The central idea of FreMo is to explicitly achieve adaptive and selective cross-modality synergy in the frequency domain.
First, to handle diverse spectral characteristics within each modality, we introduce a \emph{Modality-Wise Frequency Filter} (MFF), which adaptively refines frequency components to emphasize informative bands while suppress noise. MFF enables flexible modality-wise frequency refinement without altering temporal structure.
Second, to enable selective cross-modality synergy, we design a \emph{Frequency-Guided Synergy Integrator} (FSI), which constructs a synergy consensus based on the relative contribution of each modality at every frequency. FSI strengthens synergy in frequency components that capture shared cross-modality patterns, while limiting interference in components dominated by modality-specific variations.
FreMo is architecture-agnostic and can be integrated into existing time series models as a plug-and-play module. Our main contributions are summarized as follows:

\begin{itemize}[leftmargin=*]
    \item To the best of our knowledge, FreMo is the first work to systematically formulate frequency-domain multi-modality modeling for transportation forecasting, uncovering the effectiveness of selective and frequency-dependent synergy mechanism.
    \item We propose the MFF that performs adaptive, modality-wise frequency filtering, enabling each modality to emphasize informative components while suppressing noise.
    \item We propose the FSI that identifies and aggregates high-consensus modalities at each frequency, allowing selective and reliable cross-modality synergy.
    \item Extensive experiments demonstrate the superior performance and strong generalization of FreMo across diverse scenarios.
\end{itemize}

\section{Related Works}

\noindent\textbf{Uni-Modality Modeling.} Uni-modality time series forecasting has been widely studied.
Graph neural networks model spatial relations via diffusion-based mechanisms~\cite{li2017diffusion} , fully-connected temporal structures~\cite{oreshkin2021fc}, and cross-scale~\cite{huang2023crossgnn}, while convolutional operators capture local dependencies through spatial~\cite{deng2022graph}, temporal~\cite{wu2020connecting,wu2019graph}, spatio-temporal~\cite{guo2019deep,yang2021space,guo2019attention}, and adaptive convolutions~\cite{pan2019urban}.
\cite{jiang2021dl} provides a systematic benchmark, and \cite{zhu2021mixseq} conducts micro-cluster analysis.
Attention mechanisms~\cite{vaswani2017attention} are adopted for long-range dependencies, including spatial attention~\cite{fang2019gstnet,zheng2020gman}, temporal attention~\cite{wu2021autoformer,zhou2021informer}, adaptive embedding strategies~\cite{liu2023spatio}, with further evidence on capturing global correlations~\cite{lee2024testam,jiang2023pdformer}. 
Recent architectures integrate Transformer-based global modeling with local pattern~\cite{Yuqietal2023PatchTST,wu2023timesnet,zhang2023crossformer,liu2024itransformer,fang2025efficient}.
Self-supervised learning and large language model have been introduced to enhance time series representations~\cite{guo2021hierarchical,shao2022pre,ji2023spatio,zhang2022self,jin2023time,zhou2023one,liu2025calf}. 
Further extensions model distribution shift~\cite{fan2023dish}, heterogeneity~\cite{qiu2025duet}, lead-lag under local stationarity~\cite{zhao2024lift}, and multi-period  multi-scale transformations~\cite{wang2023timemixer,wang2025timemixer++}.
Frequency-aware modeling has attracted attention for non-stationarity and distribution shifts, including Transformer variants with frequency attention~\cite{zhou2022fedformer,cai2024msgnet}, filtering~\cite{yue2025freeformer,yi2024filternet}, debiasing ~\cite{piao2024fredformer}, frequency-domain MLPs\cite{yi2023frequency} and hypervariate graph~\cite{yi2023fouriergnn}. Related robustness studies explore frequency-adaptive normalization~\cite{ye2024frequency}, frequency-masked inference~\cite{fu2025frequency}, and time-frequency consistency pre-training \cite{zhang2022self}.
Despite their effectiveness, most existing methods focus on temporal/spatial or frequency modeling within a single modality, overlooking cross-modality interactions essential for multi-modality forecasting task.

\noindent\textbf{Multi-Modality Modeling.} Early works model cross-domain correlations using transfer learning or heterogeneous recurrent architectures across cities or demand types~\cite{yao2019learning,ye2019co}.  With the rise of graph neural networks, Zhang et al.~\cite{zhang2024multimodal} combine sparse graph attention with temporal convolution, while MDTP~\cite{fang2021mdtp} integrates GCNs and LSTMs to fuse multi-source trajectory data. Transformer-based models further capture long-range and cross-modality dependencies. MiST~\cite{huang2019mist} adopts a multi-view Transformer for complex interaction modeling, whereas DMSTGCN~\cite{han2021dynamic} fuses flow and speed representations through multi-dimensional interaction modules. Hierarchical and cross-modality Transformers explicitly model semantic dependencies across data sources. STtrans~\cite{wu2020hierarchically} introduces cross-modality attention. Several studies extend multi-modality forecasting to multi-task and multi-scenario settings by integrating heterogeneous demand signals or jointly optimizing multiple objectives~\cite{wang2022event,liu2021community,yuan2021effective}. Recent methods further address heterogeneity in multi-modality data through normalization~\cite{deng2023tts}, Gaussian mixture representation learning~\cite{deng2023gmrl}, and self-supervised learning~\cite{deng2024mossl}. Despite these advances, existing methods often rely on shallow fusion or limited cross-attention, which is insufficient for learning robust representations from noisy and sparse multi-modality data. Our method explicitly models modality reliability and selectively aggregates information from frequency domain to enable more effective cross-modality synergy.

\begin{figure*}[t]
  \centering
  \includegraphics[width=\linewidth]{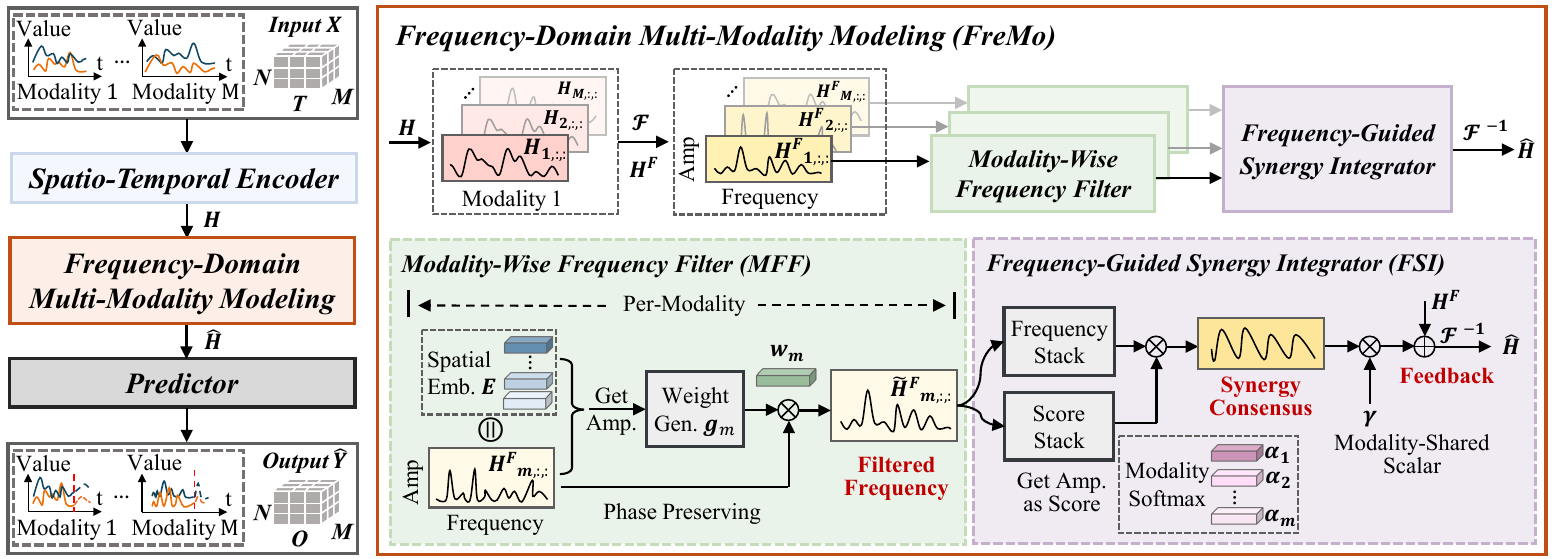}
  \caption{An overview of the proposed FreMo. Left: As a plug-and-play component, FreMo can be integrated into time series frameworks to enable cross-modality synergy. Right: Detailed design of FreMo. Modality-Wise Frequency Filter (MFF) adaptively learns modality-wise frequency weights $w$ from spectral amplitudes by incorporating spatial embedding $E$, and applies them as soft gates to preserve informative frequency components while suppressing noise in a phase-preserving manner. Frequency-Guided Synergy Integrator (FSI) computes frequency-wise synergy weights $\alpha$ across modalities, aggregates the filtered spectra into a unified synergy consensus, and injects it back through a learnable scalar $\gamma$ for residual calibration.}
  \Description{The overview of the proposed FreMo, which can be jointly trained with any general time series framework as a plug-and-play component, as illustrated on the left-hand side.}
  \label{figure:framework}
\end{figure*}

\section{Problem Definition}
Let $N$ denote the set of spatial nodes and $M$ the number of transportation modalities.
At time step $t$, the multi-modality transportation state is represented as a matrix
$X_{:,:,t} \in \mathbb{R}^{M \times N}$, where each entry corresponds to the traffic demand
of a specific modality at a given node.
Given a historical window of length $T$, the observations form a \emph{multi-modality transportation tensor}, denoted as $X = (X_{:,:,1}, \dots, X_{:,:,T}) \in \mathbb{R}^{M \times N \times T}$. Given the historical tensor $X$, the goal of multi-modality transportation forecasting is to predict traffic states for the subsequent $O$ time steps, formulated as
$\hat{Y} = ( \hat{X}_{:,:,T+1}, \dots, \hat{X}_{:,:,T+O} ) \in \mathbb{R}^{M \times N \times O}$.

\section{Methodology}
Figure~\ref{figure:framework} depicts an overview of the proposed \textbf{FreMo}. A spatio-temporal encoder initializes the latent representation from the multi-modality transportation tensor $X$. FreMo is then inserted as a plug-and-play refinement module that operates in the frequency domain and comprises two sequential components: (i) \emph{Modality-Wise Frequency Filter} (MFF), which learns modality-wise frequency weights to perform modality-adaptive, phase-preserving filtering; and (ii) \emph{Frequency-Guided Synergy Integrator} (FSI), which computes frequency-wise synergy weights across modalities, forms a shared synergy consensus, and injects it back through residual feedback. The calibrated representation is transformed back to the temporal domain and passed to a predictor to generate future forecasts.

\subsection{Spatio-Temporal Encoder}
In this study, we design a spatio-temporal encoder to jointly preserve spatio-temporal contextual information in multi-modality transportation data. The encoder aims to unifiedly capture sequential patterns across time periods, spatial dependencies among nodes, and shared representations across modalities. For notational simplicity, a nonlinear projection function is defined as $f(x)=ReLU(x \cdot W+b)$, where $W$ and $b$ are trainable parameters. The Encoder takes the projected representation $H^{(\text{in})}=f(X) \in \mathbb{R}^{M \times N \times T \times d}$ as input.

\noindent \textbf{Temporal Encoder.} 
We adopt a Temporal Convolution Network (TCN)~\cite{Oord2016WaveNetAG} as the Temporal Encoder to fuse high-dimensional temporal features. The architecture comprises two dilated causal convolutions to enlarge the receptive field for learned representation, a gated linear unit for incorporating nonlinearity into the network, and a 1 × 1 convolution operation for refining features. The Temporal Encoder is defined as follows:
\begin{equation}
H^{(\text{te})} = \text{Conv}\left(\text{tanh}(H^{(\text{in})} \ast W_{\text{f}}) \odot \sigma (H^{(\text{in})} \ast W_{\text{g}})\right),
\end{equation}
\noindent where $\ast$ denotes the dilated causal convolution, $W_{\text{f}}$ and $W_{\text{g}}$ are learnable weights for the filter and gate, respectively, and $\odot$ indicates element-wise multiplication. The operator $\text{Conv}(\cdot)$ represents a 1$\times$1 convolution.

\noindent \textbf{Spatial Encoder.} 
To capture spatial dependencies among nodes while circumventing the quadratic complexity of standard self-attention (i.e., $\mathcal{O}(N^2)$), we introduce a latent-based spatial encoding mechanism motivated by~\cite{lee2019set}. We initialize a set of learnable latent variables $Z \in \mathbb{R}^{L \times d} (L \ll d)$ to act as a global information bottleneck. The Spatial Encoder consists of two cross-attention stages with Layer Normalization \cite{ba2016layer} followed by a feed-forward network (FFN). First, the latent variables query the node-level information to aggregate global contexts:
\begin{equation} 
\tilde{Z} = \text{LN}\left(Z + \text{CrossAttn}(Z, H^{(\text{te})}, H^{(\text{te})})\right),
\end{equation}
\noindent where $\text{LN}(\cdot)$ denotes Layer Normalization, and $\text{CrossAttn}(Q,K,V)$ represents the cross-attention mechanism. 
Then, the updated latents $\tilde{Z}$ act as keys and values to refine the node representations, distributing the global information back to the spatial units:
\begin{equation} 
\begin{aligned}
H^{(\text{se})} &= \text{LN}\left(H^{(\text{te})} + \text{CrossAttn}(H^{(\text{te})}, \tilde{Z}, \tilde{Z})\right), \\
H^{(\text{se})} &\leftarrow H^{(\text{se})} + \text{FFN}(H^{(\text{se})}).
\end{aligned}
\end{equation}

The resulting representation $H^{(\text{se})}$ preserves the original multi-modality spatio-temporal structure while incorporating latent-guided spatial interactions with a linear complexity $\mathcal{O}(LN)$.

A stacked architecture comprising multiple layers is employed to model spatio-temporal interactions across modalities effectively. In each layer, the input undergoes a Temporal Encoder followed by a Spatial Encoder via residual learning, transforming the raw data into a comprehensive latent representation $H \in \mathbb{R}^{M \times N \times T \times d}$.

\subsection{Modality-Wise Frequency Filter}
Multi-modality transportation signals exhibit frequency-dependent structure, where informative components and noise distribute differently across frequencies. Modalities also have distinct spectral profiles: some are dominated by low-frequency periodic trends, whereas others contain pronounced high-frequency fluctuations. Moreover, informative frequencies may vary across spatial units. These factors make a single shared frequency filter suboptimal: it may over-suppress useful components for certain modalities/nodes, or retain excessive noise for others. 
To address this challenge, we propose the \emph{Modality-Wise Frequency Filter} (MFF), which performs adaptive and independent frequency filtering for each modality and node. MFF derives modality-wise frequency gates from spectral amplitude profiles and applies them in a phase-preserving manner, so that informative frequency components are emphasized while noise is suppressed without distorting temporal alignment. 

\noindent \textbf{Frequency-Domain Transformation.}
We first expose frequency components explicitly by transforming temporal representations into the frequency domain via real-valued FFT (rFFT):
\begin{equation} 
H^{\text{F}}_{m,n,:} = \mathcal{F}(H_{m,n,:}) \in \mathbb{C}^{F \times d}, 
\end{equation}
\noindent where $\mathcal{F}(\cdot)$ denotes the rFFT operator and $F = \left\lfloor \frac{T}{2} \right\rfloor + 1$ is the number of frequency bins. We then compute the amplitude spectrum as:
\begin{equation} 
A_{m,n,:} = |H^{\text{F}}_{m,n,:}| \in \mathbb{R}^{F \times d},
\end{equation}
\noindent where $|\cdot|$ denotes the modulus of the complex number.

\noindent \textbf{Modality-Wise Frequency Weight Generation.}
The amplitude spectrum provides a direct cue of how strong each frequency component is. Based on $A_{m,n,:}$, we learn a frequency-wise gating vector that highlights informative spectral bands while attenuating noise. To capture both modality differences and spatial variation, we use an independent weight generator for each modality and incorporate a learnable node embedding as local context. 
Specifically, for the $m^\text{th}$ modality, we first aggregate the amplitude spectrum along the channel dimension to obtain $\bar{A}_{m,n,:} \in \mathbb{R}^{F}$, and then concatenate it with the node embedding $E_n \in \mathbb{R}^{L}$  to form the context $\phi_{m,n,:} \in \mathbb{R}^{F + L}$:
\begin{equation} 
\begin{aligned} 
\bar{A}_{m,n,:} &= \text{Pool}_d(A_{m,n,:}), \\
\phi_{m,n,:} &= [\bar{A}_{m,n,:} \parallel E_n ], 
\end{aligned} 
\end{equation}
\noindent where $\text{Pool}_d(\cdot)$ denotes average pooling over the hidden dimension and $\parallel$ represents concatenation. The modality-wise weight generator $\mathcal{G}_m(\cdot)$ maps the context to a frequency-bin gate:
\begin{equation} 
w_{m,n,:} = \sigma\left(\mathcal{G}_m(\phi_{m,n,:})\right) \in [0, 1]^{F},
\label{eq:fre_weight}
\end{equation}
\noindent where $\sigma(\cdot)$ is the Sigmoid activation function. The resulting $w_{m,n,:}$ is a modality-wise frequency weight for modality $m$ at node $n$, which assigns a retention score to each frequency bin: larger values keep the corresponding components, while smaller values attenuate them, thereby serving as a soft frequency gate within the modality.

\noindent \textbf{Phase-Preserving Frequency Filtering.}
Based on the learned gate $w_{m,n,:}$, we then modulate the complex spectrum to perform modality-wise frequency filtering:
\begin{equation} 
\tilde{H}^{\text{F}}_{m,n,:} = H^{\text{F}}_{m,n,:} \odot w_{m,n,:},
\end{equation}
\noindent where $\odot$ denotes the element-wise Hadamard product with broadcasting over the hidden dimension. Since $w_{m,n,:}$ is real-valued, this operation rescales amplitudes while preserving phase, thereby suppressing noise without disrupting temporal alignment. After obtaining filtered frequency representations of each modality independently, we gather the results as $\tilde{H}^{\text{F}} \in \mathbb{C}^{M \times N \times F \times d}$.

\subsection{Frequency-Guided Synergy Integrator}
While MFF independently refines spectral components for each modality, effective forecasting also depends on cross-modality synergy. Synergy is inherently frequency-dependent: low-frequency trends (e.g., daily commuting) often consistent across modalities, whereas high-frequency components are more likely to be modality-specific details. Existing fusion methods \cite{huang2019mist,deng2023gmrl,zhang2024multimodal,fang2021mdtp,han2021dynamic} typically operate in the time domain or adopt shallow interactions, and thus do not explicitly model frequency-wise modality consensus. To bridge this gap, we propose the Frequency-Guided Synergy Integrator (FSI). FSI constructs a frequency-wise synergy consensus by assigning a relative modality weight at each frequency, and then uses consensus to calibrate modality-wise spectral representations.

\noindent \textbf{Frequency-Wise Synergy Weighting.}
To quantify how much each modality should contribute at each frequency bin, we compute a frequency-wise strength score from the filtered spectrum $\tilde{H}^{\text{F}}_{m,n,:}$:
\begin{equation} 
S_{m,n,:} = \text{Pool}_d(|\tilde{H}^{\text{F}}_{m,n,:}|), 
\end{equation}
\noindent where $S_{m,n,:} \in \mathbb{R}^{F}$ serves as a proxy of spectral strength for modality $m$ at node $n$, and larger values indicate stronger responses at the corresponding frequencies. We then apply the Softmax over modalities to obtain the frequency-wise synergy weights:
\begin{equation} 
\alpha_{m,n,:} = \frac{\text{exp}(S_{m,n,:})}{\sum\limits_{m^{\prime} \in M} \text{exp}(S_{m^{\prime},n,:})}, 
\label{eq:score}
\end{equation}
\noindent where $\alpha_{m,n,:} \in \mathbb{R}^{F}$ gives the relative contribution of $m^\text{th}$ modality at each frequency bin (with $\sum_m \alpha_{m,n,f}=1$), enabling frequency-granular competition and supporting subsequent synergy consensus construction. Large $\alpha_{m,n,f}$ indicates that modality $m$ is more reliable for forming the cross-modality consensus at frequency $f$.

\noindent \textbf{Synergy Consensus Construction.}
Given the learned frequency-wise synergy weights $\alpha_{m,n,:}$, we form a unified \emph{synergy consensus} by aggregating the filtered spectra of all modalities with frequency-granular weighting:
\begin{equation} 
\mathcal{C}_{n,:} = \sum_{m=1}^{M} \alpha_{m,n,:} \odot \tilde{H}^{\text{F}}_{m,n,:}, 
\end{equation}
\noindent where $\odot$ denotes element-wise multiplication with broadcasting over the hidden dimension. The resulting $\mathcal{C}_{n,:} \in \mathbb{C}^{F \times d}$ summarizes the cross-modality consensus at each frequency bin, serving as a shared frequency-domain reference for all modalities.

\noindent \textbf{Feedback Injection and Reconstruction.}
The synergy consensus is then fed back to each modality via residual injection to calibrate its original spectrum:
\begin{equation} 
\hat{H}^{\text{F}}_{m,n,:} = H^{\text{F}}_{m,n,:} + \gamma \cdot \mathcal{C}_{n,:}, 
\end{equation}
\noindent where $\gamma \in \mathbb{R}$ is a learnable scalar initialized to zero and shared across modalities. This global scaling factor regulates the strength of synergy feedback, balancing modality-specific spectra with the shared consensus. Finally, the calibrated frequency representation is transformed back to the temporal domain:
\begin{equation} 
\hat{H}_{m,n,:} = \mathcal{F}^{-1}(\hat{H}^{\text{F}}_{m,n,:}). 
\end{equation}
\noindent where $\mathcal{F}^{-1}(\cdot)$ denotes inverse rFFT, yielding the augmented temporal representation $\hat{H} \in \mathbb{R}^{M \times N \times T \times d}$. The complete workflow of FreMo is summarized in Algorithm~\ref{alg:fremo}.

\subsection{Model Optimization}
Given the refined multi-modality representation obtained as described above, we employ a Predictor to project it into the future horizon. Specifically, our Predictor contains a gated linear unit and a linear projection layer, which is formulated as follows:
\begin{equation} 
\hat{Y}_{m,n,:} = \left((\hat{H}_{m,n,:} \ast W_\text{vc}) \odot \sigma(\hat{H}_{m,n,:} \ast W_\text{gc})\right) \ast W_{\text{o}},
\end{equation}
\noindent where $W_\text{vc}, W_\text{gc} \in \mathbb{R}^{T \times d \times d}$ denote temporal convolution kernels, and $W_{\text{o}} \in \mathbb{R}^{d \times O}$ projects the gated features to the next $O$-step horizon.

The model is trained in an end-to-end manner by minimizing the Mean Absolute Error (MAE) between the predicted multi-modality transportation tensor $\hat{Y} \in \mathbb{R}^{M \times N \times O}$ and the ground truth $Y$:
\begin{equation}
\mathcal{L} = \frac{1}{M \times N \times O} \sum_{m=1}^{M} \sum_{n=1}^{N} \sum_{t=1}^{O} | \hat{Y}_{m,n,t} - Y_{m,n,t} |.
\end{equation}

\begin{algorithm}[t]
\caption{Pseudocode of the proposed FreMo}
\label{alg:fremo}
\begin{algorithmic}[1]
\REQUIRE 
    Input hidden representation $\mathbf{H} \in \mathbb{R}^{M \times  N \times T \times d}$; \\
    Learnable node embeddings $\mathbf{E} \in \mathbb{R}^{N \times l}$; \\
    Modality-wise weight generators $\{g_m(\cdot)\}_{m=1}^M$; \\
    Modality-shared learnable scalar $\gamma$.
\ENSURE 
    Refined hidden representation $\mathbf{\hat{H}}$.

\STATE \textbf{// Step 1: Frequency Domain Transformation}
\STATE $\mathbf{H}^F \leftarrow \text{rFFT}(\mathbf{H}, \text{dim}=T)$ \hfill $\rhd$ Transform to frequency domain
\STATE $\mathbf{A} \leftarrow |\mathbf{H}^F|$ \hfill $\rhd$ Compute spectral amplitude

\STATE \textbf{// Step 2: Modality-Wise Frequency Filter (MFF)}
\FOR{$m = 1$ to $M$}
    \STATE Extract amplitude $\mathbf{A}_m$ and complex features $\mathbf{H}^F_m$ for modality $m$
    \STATE $\mathbf{\bar{A}}_m \leftarrow \text{Mean}(\mathbf{A}_m, \text{dim}=d)$ \hfill $\rhd$ Aggregate hidden dimension
    \STATE $\mathbf{\phi}_m \leftarrow \text{Concat}(\mathbf{\bar{A}}_m, \mathbf{E})$ \hfill $\rhd$ Inject node embedding
    \STATE $\mathbf{w}_m \leftarrow g_m(\mathbf{\phi}_m)$ \hfill $\rhd$ Generate frequency weights $\in [0, 1]$
    \STATE $\tilde{\mathbf{H}}^F_m \leftarrow \mathbf{H}^F_m \odot \mathbf{w}_m$ \hfill $\rhd$ Phase-preserving filtering
    \STATE $\mathbf{S}_m \leftarrow \text{Mean}(|\tilde{\mathbf{H}}^F_m|, \text{dim}=d)$ \hfill $\rhd$ Compute reliability score
\ENDFOR

\STATE \textbf{// Step 3: Frequency-Guided Synergy Integrator (FSI)}
\STATE $\boldsymbol{\alpha} \leftarrow \text{Softmax}(\text{Stack}([\mathbf{S}_1, \dots, \mathbf{S}_M]), \text{dim}=M)$ \hfill $\rhd$ Calculate frequency-wise synergy weight
\STATE $\mathcal{C}^F \leftarrow \sum_{m=1}^M \alpha_m \odot \tilde{\mathbf{H}}^F_m$ \hfill $\rhd$ Construct synergy consensus

\STATE $\mathbf{\hat{H}}^F \leftarrow \mathbf{H}^F + \gamma \cdot \mathcal{C}^F$ \hfill $\rhd$ Inject consensus via residual connection
\STATE $\mathbf{\hat{H}} \leftarrow \text{irFFT}(\mathbf{\hat{H}}^F, \text{dim}=F)$ \hfill $\rhd$ Reconstruct to time domain

\RETURN $\mathbf{\hat{H}}$
\end{algorithmic}
\end{algorithm}

\section{Experiment}
\subsection{Experiment Setup}
\subsubsection{Datasets.}
\begin{table}[H]
\centering
\caption{Summary of multi-modality transportation datasets.}
\label{tab:datasets}
\renewcommand{\arraystretch}{0.9}
\resizebox{\columnwidth}{!}{%
\begin{tabular}{l | l | c | c | c}
\toprule \noalign{\smallskip}
\textbf{Dataset} & \textbf{Time Period} & \textbf{Nodes} & \textbf{Modalities} & \textbf{Horizons} \\ \noalign{\smallskip} \hline \noalign{\smallskip}

NYC & \begin{tabular}[c]{@{}l@{}}2016/4/1 $\sim$ \\ 2016/6/30 (0.5h)\end{tabular} & 98 & \multirow{6.5}{*}{ \begin{tabular}[c]{@{}c@{}} Bike Inflow, \\ Bike Outflow, \\ Taxi Inflow, \\ Taxi Outflow \end{tabular} } & \multirow{6.5}{*}{ \begin{tabular}[c]{@{}c@{}} Input: 16 \\ $\downarrow$ \\ Out: 1/2/3 \end{tabular} } \\ 
\noalign{\smallskip} \noalign{\smallskip}

DC & \begin{tabular}[c]{@{}l@{}}2015/10/24 $\sim$ \\ 2016/1/31 (1h)\end{tabular} & 108 &  &  \\ 
\noalign{\smallskip} \noalign{\smallskip}

Chicago & \begin{tabular}[c]{@{}l@{}}2016/4/1 $\sim$ \\ 2016/6/30 (0.5h)\end{tabular} & 510 &  &  \\ \noalign{\smallskip}
\bottomrule
\end{tabular}
}
\end{table}

\begin{table*}[t]
\centering
\scriptsize
\caption{Performance on multi-modality transportation datasets for the first/second/third horizon. The best accuracy is highlighted in \textbf{bold}, and the second-best performance is \underline{underlined}.}
\renewcommand{\arraystretch}{1}
\setlength{\tabcolsep}{2.5pt}
\resizebox{\textwidth}{!}{%
    \begin{tabular}{c|cc|cc|cc|cc}
    \toprule    
    \toprule
    \multirow{2}{*}{\textbf{NYC}} & 
    \multicolumn{2}{c|}{Bike Inflow} & 
    \multicolumn{2}{c|}{Bike Outflow} & 
    \multicolumn{2}{c|}{Taxi Inflow} & 
    \multicolumn{2}{c}{Taxi Outflow} \\
    \cmidrule(lr){2-3} \cmidrule(lr){4-5} \cmidrule(lr){6-7} \cmidrule(lr){8-9}
    
     & MAE & RMSE & MAE & RMSE & MAE & RMSE & MAE & RMSE \\
    \midrule
    
    (GWN)~\cite{wu2019graph} &
    2.37 / 2.66 / 2.91 & 6.89 / 7.63 / 8.23 & 2.57 / 2.91 / 3.17 & 7.42 / 8.26 / 8.85 & 7.00 / 8.23 / 9.59 & 13.30 / 16.11 / 18.99 & 7.72 / 9.05 / 10.29 & 14.98 / 17.48 / 19.69 \\
    
    AGCRN~\cite{bai2020adaptive} &
    2.40 / 2.70 / 3.02 & 6.96 / 7.69 / 8.41 & 2.60 / 2.97 / 3.31 & 7.39 / 8.32 / 9.15 & 7.08 / 8.10 / 9.16 & 13.42 / 15.71 / 18.16 & 7.81 / 9.03 / 10.19 & 15.03 / 17.41 / 19.29 \\
    
    MTGNN~\cite{wu2020connecting} &
    2.28 / 2.44 / 2.69 & 6.68 / 7.05 / 7.62 & 2.45 / 2.60 / 2.83 & 7.07 / 7.46 / 8.02 & 7.24 / 8.04 / 8.78 & 13.82 / 15.82 / 17.71 & 7.97 / 8.97 / 9.96 & 15.47 / 17.59 / 19.26 \\
    
    TimesNet~\cite{wu2023timesnet} &
    2.65 / 2.88 / 3.21 & 4.61 / 5.11 / 5.88  & 2.81 / 3.10 / 3.45 & 5.03 / 5.65 / 6.34 & 7.54 / 8.44 / 9.25 & 13.61 / 15.98 / 18.04 & 8.45 / 9.52 / 10.57 & 15.94 / 18.15 / 20.37 \\
    
    iTransformer~\cite{liu2024itransformer} &
    3.05 / 3.75 / 4.52 & 5.61 / 7.22 / 8.85 & 3.31 / 4.10 / 4.80 & 6.20 / 7.99 / 9.44 & 8.11 / 10.32 / 12.54 & 14.40 / 19.16 / 23.98 & 8.69 / 10.85 / 12.80 & 16.15 / 20.60 / 24.37 \\
    
    STAEformer~\cite{liu2023spatio} & 3.57 / 3.67 / 3.81 & 7.13 / 7.42 / 7.78 & 3.44 / 3.57 / 3.73 & 7.10 / 7.40 / 7.74 & 8.76 / 10.33 / 11.89 & 15.40 / 19.51 / 24.06 & 9.68 / 10.96 / 12.14 & 16.74 / 19.34 / 21.75 \\
    
    MiST~\cite{huang2019mist} &
    2.32 / 2.52 / 2.76 & 6.78 / 7.26 / 7.84 & 2.57 / 2.83 / 3.01 & 7.48 / 8.10 / 8.54 & 7.93 / 8.73 / 9.61 & 14.10 / 16.51 / 19.06 & 8.44 / 9.19 / 10.82 & 15.56 / 17.93 / 19.33 \\
    
    STtrans~\cite{wu2020hierarchically} &
    2.72 / 3.06 / 3.38 & 7.74 / 8.54 / 9.35 & 2.96 / 3.27 / 3.56 & 8.33 / 8.93 / 9.67 & 8.46 / 9.62 / 11.03 & 15.67 / 17.74 / 20.42 & 8.78 / 10.04 / 11.28 & 16.66 / 19.12 / 21.39 \\
    
    COCOA~\cite{deldari2022cocoa} &
    2.82 / 3.11 / 3.48 & 5.06 / 5.82 / 6.53 & 2.93 / 3.28 / 3.60 & 5.43 / 6.23 / 6.85 & 8.05 / 9.81 / 11.95 & 13.98 / 17.85 / 22.18 & 8.80 / 10.69 / 12.83 & 16.20 / 20.15 / 24.02 \\
    
    MoSSL~\cite{deng2024mossl} &
    \underline{2.26} / \underline{2.39} / \underline{2.58} & \underline{4.10} / \underline{4.28} / \underline{4.51} & \underline{2.43} / \underline{2.55} / \underline{2.67} & \underline{4.42} / \underline{4.65} / \underline{4.86} & \underline{6.73} / \underline{7.53} / \underline{8.38} & \underline{11.89} / \underline{14.13} / \underline{16.22} & \underline{7.37} / \underline{8.18} / \underline{8.90} & \underline{13.80} / \underline{15.61} / \underline{17.23} \\
    
    \textbf{FreMo (Ours)} & 
    \textbf{2.22 / 2.33 / 2.42} & \textbf{4.06 / 4.21 / 4.39} & \textbf{2.34 / 2.45 / 2.55} & \textbf{4.40 / 4.61 / 4.80} & \textbf{6.43 / 7.01 / 7.54} & \textbf{11.83 / 13.85 / 15.78} & \textbf{7.07 / 7.67 / 8.25} & \textbf{13.49 / 14.71 / 15.86} \\
    \midrule
    \midrule

    \multirow{2}{*}{\textbf{DC}} & 
    \multicolumn{2}{c}{Bike Inflow} & 
    \multicolumn{2}{c}{Bike Outflow} & 
    \multicolumn{2}{c}{Taxi Inflow} & 
    \multicolumn{2}{c}{Taxi Outflow} \\    
    \cmidrule(lr){2-3} \cmidrule(lr){4-5} \cmidrule(lr){6-7} \cmidrule(lr){8-9}
    
     & MAE & RMSE & MAE & RMSE & MAE & RMSE & MAE & RMSE \\
    \midrule
    
    (GWN) \cite{wu2019graph} & 
    0.87 / 0.93 / 0.99 & 1.51 / 1.60 / 1.66 & 0.88 / 0.94 / 1.00 & 1.53 / 1.62 / 1.67 & 3.59 / 3.78 / 4.00 & 6.57 / 6.85 / 7.21 & 4.03 / 4.22 / 4.39 & 8.55 / 8.87 / 9.19 \\
    
    AGCRN~\cite{bai2020adaptive} &
    0.99 / 1.10 / 1.24 & 1.49 / 1.63 / 1.79 & 1.00 / 1.11 / 1.25 & 1.51 / 1.65 / 1.81 & 2.76 / 3.00 / 3.29 & 4.69 / 5.22 / 5.68 & 2.95 / 3.29 / 3.61 & 5.73 / 6.65 / 7.32 \\
        
    MTGNN~\cite{wu2020connecting} &
    0.67 / 0.75 / 0.81 & 1.23 / 1.38 / 1.48 & 0.69 / 0.77 / 0.82 & 1.27 / 1.40 / 1.50 & 2.78 / 3.12 / 3.38 & 4.90 / 5.73 / 6.21 & 2.95 / 3.41 / 3.68 & 6.04 / 7.23 / 7.83 \\
    
    TimesNet~\cite{wu2023timesnet} &
    \underline{0.54} / \underline{0.59} / \underline{0.62} & 1.17 / 1.27 / 1.32 & \underline{0.56} / \underline{0.60} / \underline{0.63} & \underline{1.20} / 1.30 / 1.35 & 2.55 / 2.78 / 2.94 & \underline{4.38} / \textbf{4.87} / 5.58 & 2.79 / 3.04 / \underline{3.22} & \underline{5.41} / \underline{5.92} / 7.26 \\
    
    iTransformer~\cite{liu2024itransformer} &
    0.63 / 0.71 / 0.77 & 1.38 / 1.55 / 1.65 & 0.64 / 0.71 / 0.77 & 1.41 / 1.55 / 1.66 & 3.10 / 3.70 / 4.24 & 5.35 / 6.60 / 7.58 & 3.26 / 4.03 / 4.60 & 6.37 / 8.08 / 9.33 \\
    
    STAEformer~\cite{liu2023spatio} & 
    0.64 / 0.71 / 0.73 & 1.23 / 1.31 / 1.35 & 0.65 / 0.71 / 0.75 & 1.28 / 1.34 / 1.38 & 2.94 / 3.31 / 3.70 & 5.36 / 6.23 / 6.95 & 3.36 / 3.52 / 3.81 & 7.71 / 7.94 / 8.51 \\
    
    MiST~\cite{huang2019mist} &
    0.72 / 0.80 / 0.91 & 1.37 / 1.55 / 1.68 & 0.74 / 0.81 / 0.92 & 1.43 / 1.59 / 1.71 & 3.05 / 3.64 / 4.09 & 5.39 / 6.61 / 7.48 & 3.25 / 3.87 / 4.37 & 7.00 / 8.49 / 9.56 \\
    
    STtrans~\cite{wu2020hierarchically} &
    0.68 / 0.81 / 0.87 & 2.09 / 2.42 / 2.67 & 0.74 / 0.83 / 0.89 & 2.26 / 2.43 / 2.63 & 3.13 / 3.51 / 3.81 & 5.85 / 6.54 / 6.95 & 3.21 / 3.66 / 4.26 & 6.56 / 7.71 / 8.51 \\
    
    COCOA~\cite{deldari2022cocoa} & 0.70 / 0.71 / 0.79 & 1.28 / 1.36 / 1.45 & 0.71 / 0.74 / 0.80 & 1.30 / 1.38 / 1.47 & 2.68 / 3.03 / 3.31 & 4.77 / 5.64 / 6.23 & 2.88 / 3.37 / 3.70 & 5.89 / 7.08 / 7.94 \\
    
    MoSSL~\cite{deng2024mossl} &
    0.56 / 0.63 / 0.66 & \underline{1.11} / \underline{1.21} / \underline{1.29} & 0.61 / 0.65 / 0.67 & \underline{1.20} / \underline{1.27} / \underline{1.32} & \underline{2.49} / \underline{2.75} / \underline{2.89} & 4.45 / 5.21 / \underline{5.51} & \underline{2.69} / \underline{2.98} / 3.36 & 5.44 / 6.63 / \underline{6.87} \\
    
    \textbf{FreMo (Ours)} &
    \textbf{0.46 / 0.51 / 0.54} & \textbf{1.07 / 1.18 / 1.21} & \textbf{0.50 / 0.52 / 0.55} & \textbf{1.16 / 1.20 / 1.26} & \textbf{2.39 / 2.60 / 2.82} & \textbf{4.33} / \underline{4.94} / \textbf{5.40} & \textbf{2.57 / 2.90 / 3.11} & \textbf{5.33 / 6.29 / 6.61} \\
    \midrule
    \midrule

    \multirow{2}{*}{\textbf{Chicago}} & 
    \multicolumn{2}{c}{Bike Inflow} & 
    \multicolumn{2}{c}{Bike Outflow} & 
    \multicolumn{2}{c}{Taxi Inflow} & 
    \multicolumn{2}{c}{Taxi Outflow} \\    
    \cmidrule(lr){2-3} \cmidrule(lr){4-5} \cmidrule(lr){6-7} \cmidrule(lr){8-9}
    
     & MAE & RMSE & MAE & RMSE & MAE & RMSE & MAE & RMSE \\
    \midrule
    
    (GWN)~\cite{wu2019graph} & 
    0.43 / 0.45 / 0.48 & 1.79 / 1.90 / 2.03 & 0.43 / 0.46 / 0.49 & 1.88 / 1.99 / 2.08 & 0.75 / 0.81 / 0.86 & 3.66 / 3.97 / 4.31 & 0.68 / 0.74 / 0.79 & 3.73 / 4.11 / 4.48 \\
    
    AGCRN~\cite{bai2020adaptive} & 
    0.51 / 0.52 / 0.60 & 1.35 / 1.52 / 1.75 & 0.52 / 0.53 / 0.60 & 1.45 / 1.57 / 1.73 & 0.76 / 0.79 / 0.90 & 2.51 / 2.74 / 3.18 & 0.69 / 0.73 / 0.85 & 2.53 / 2.87 / 3.44 \\
    
    MTGNN~\cite{wu2020connecting} &
    0.32 / 0.34 / 0.35 & \underline{1.19} / 1.27 / 1.37 & 0.33 / 0.34 / 0.35 & 1.28 / 1.34 / 1.42 & 0.56 / 0.60 / 0.65 & 2.33 / 2.54 / 2.89 & 0.50 / 0.54 / 0.61 & 2.40 / 2.73 / 3.29 \\
    
    TimesNet~\cite{wu2023timesnet} &
    0.34 / 0.36 / 0.39 & 1.25 / 1.38 / 1.56 & 0.35 / 0.37 / 0.39 & 1.35 / 1.43 / 1.54 & 0.59 / 0.64 / 0.69 & 2.56 / 2.88 / 3.26 & 0.52 / 0.57 / 0.63 & 2.66 / 2.99 / 3.54 \\
    
    iTransformer~\cite{liu2024itransformer} &
    0.38 / 0.43 / 0.49 & 1.46 / 1.76 / 2.07 & 0.38 / 0.41 / 0.45 & 1.54 / 1.72 / 1.87 & 0.61 / 0.72 / 0.84 & 2.60 / 3.16 / 3.90 & 0.55 / 0.65 / 0.79 & 2.63 / 3.29 / 4.22 \\
    
    STAEformer~\cite{liu2023spatio} & 
    0.35 / 0.35 / 0.37 & 1.91 / 1.92 / 1.96 & 0.37 / 0.38 / 0.40 & 1.98 / 2.01 / 2.07 & 0.74 / 0.78 / 0.83 & 4.65 / 4.93 / 5.20 & 0.95 / 0.99 / 1.03 & 4.68 / 4.87 / 5.06 \\
    
    MiST~\cite{huang2019mist} &
    0.34 / 0.38 / 0.43 & 1.35 / 1.55 / 1.76 & 0.35 / 0.39 / 0.43 & 1.42 / 1.57 / 1.74 & 0.57 / 0.64 / 0.72 & 2.42 / 2.75 / 3.28 & 0.51 / 0.57 / 0.65 & 2.45 / 2.79 / 3.32 \\
    
    STtrans~\cite{wu2020hierarchically} &
    0.37 / 0.41 / 0.45 & 2.26 / 2.54 / 2.74 & 0.39 / 0.42 / 0.45 & 2.28 / 2.35 / 2.73 & 0.68 / 0.72 / 0.78 & 3.02 / 3.07 / 3.37 & 0.56 / 0.60 / 0.70 & 2.80 / 3.19 / 3.78 \\
    
    COCOA~\cite{deldari2022cocoa} & 
    0.34 / 0.35 / 0.38 & 1.30 / 1.45 / 1.63 & 0.34 / 0.36 / 0.38 & 1.35 / 1.45 / 1.58 & 0.60 / 0.67 / 0.77 & 2.60 / 2.94 / 3.57 & 0.54 / 0.63 / 0.74 & 2.69 / 3.29 / 4.16 \\
    
    MoSSL~\cite{deng2024mossl} &
    \underline{0.31} / \underline{0.32} / \underline{0.34} & 1.20 / \underline{1.26} / \underline{1.31} & \underline{0.32} / \underline{0.33} / \underline{0.34} & \underline{1.21} / \underline{1.24} / \underline{1.31} & \underline{0.54} / \underline{0.56} / \underline{0.60} & \underline{2.29} / \underline{2.45} / \underline{2.72} & \underline{0.47} / \underline{0.49} / \underline{0.55} & \underline{2.32} / \underline{2.61} / \underline{3.05} \\
    
    \textbf{FreMo (Ours)} &
    \textbf{0.28 / 0.29 / 0.30} & \textbf{1.15 / 1.20 / 1.25} & \textbf{0.28 / 0.29 / 0.30} & \textbf{1.17 / 1.20 / 1.24} & \textbf{0.50 / 0.52 / 0.56} & \textbf{2.21 / 2.37 / 2.63} & \textbf{0.44 / 0.46 / 0.50} & \textbf{2.22 / 2.42 / 2.70} \\
    \bottomrule
    \bottomrule
    \end{tabular}%
}
\label{tab:performance}
\end{table*}

To fully evaluate the proposed FreMo under different scenarios, we conduct experiments on three real-world multi-modality transportation datasets, as presented in Table~\ref{tab:datasets}. These city-level datasets, collected from New York City (NYC), Washington D.C. (DC), and Chicago, span distinct spatio-temporal scales and coverage. Specifically, multi-modality is represented by four distinct transport modes, including Bike Inflow, Bike Outflow, Taxi Inflow, and Taxi Outflow.

\subsubsection{Implementation Details.}
For the model, the spatio-temporal encoder consists of four layers with a hidden dimension of 64. The temporal encoder adopts a kernel size of 3, while the spatial encoder employs a latent dimension of 64. The node embedding dimension in FreMo is set the same as the latent dimension. The training phase is performed using the Adam optimizer, and the batch size is 16. The inputs are normalized by Z-Score.
We implement FreMo using PyTorch, and conduct all experiments on a GPU server with NVIDIA GeForce GTX 2080 Ti graphic cards. For evaluation, we adopt Mean Absolute Error (MAE) and Root Mean Square Error (RMSE) as evaluation metrics.

\subsubsection{Baselines.}
We choose ten state-of-the-art models as baselines, which can be broadly categorized into two groups. (1) \textbf{Uni-modality forecasting models:} well-established general time series forecasting models, including Graph WaveNet (GWN) \cite{wu2019graph}, AGCRN \cite{bai2020adaptive}, MTGNN \cite{wu2020connecting}, TimesNet \cite{wu2023timesnet}, iTransformer \cite{liu2024itransformer}, and STAEformer \cite{liu2023spatio}. (2) \textbf{Multi-modality forecasting models:} representative methods explicitly designed for multi-modality modeling, including MiST \cite{huang2019mist}, STtrans \cite{wu2020hierarchically}, COCOA \cite{deldari2022cocoa}, and MoSSL \cite{deng2024mossl}.

\subsection{Overall Performance}
The average results of three repeated experiments are listed in Table~\ref{tab:performance}, with the best in bold and the second underlined.
Among the uni-modality baselines, MTGNN and TimesNet consistently outperform other models. MTGNN benefits from graph-based temporal dependency modeling that partially captures structured spatial correlations within each modality. TimesNet further demonstrates competitive performance, achieving second-best results on a substantial portion of metrics in the Washington DC dataset, owing to its frequency-aware period discovery mechanism. Nevertheless, these models remain inherently limited in multi-modality settings, as they process each modality independently and fail to exploit complementary cross-modality information, resulting in suboptimal performance when inter-modality dynamics are critical.
In contrast, most multi-modality baselines (i.e., MiST, STtrans, and COCOA) even underperform strong uni-modality models, primarily due to coarse or static fusion strategies that indiscriminately aggregate modalities without considering their varying reliability across temporal patterns. MoSSL stands out by achieving second-best performance across all datasets, benefiting from self-supervised objectives that promote selective cross-modality alignment and alleviate negative transfer. However, MoSSL operates mainly in the temporal domain and lacks explicit mechanisms to regulate modality contributions at finer structural levels. FreMo consistently achieves the best performance by explicitly modeling modality-wise spectral patterns and coordinating cross-modality collaboration in the frequency domain, enabling fine-grained signal selection and robust information sharing.

\begin{table}[t]
\centering
\caption{Ablation studies on the NYC. Results are taken from the third horizon.}
\setlength{\tabcolsep}{1.5pt}
\resizebox{\linewidth}{!}{%
    \begin{tabular}{c|cc|cc|cc|cc}
    \toprule    
    \multirow{2}{*}{Variant} & 
    \multicolumn{2}{c|}{Bike Inflow} & 
    \multicolumn{2}{c|}{Bike Outflow} & 
    \multicolumn{2}{c|}{Taxi Inflow} & 
    \multicolumn{2}{c}{Taxi Outflow} \\
    
    \cmidrule(lr){2-3} \cmidrule(lr){4-5} \cmidrule(lr){6-7} \cmidrule(lr){8-9}

     & MAE & RMSE & MAE & RMSE & MAE & RMSE & MAE & RMSE \\
    \midrule
    
    w/o FreMo &
    2.64 & 4.91 & 2.67 & 5.19 & 8.80 & 17.60 & 9.15 & 17.71 \\

    Time Domain &
    2.52 & 4.61 & 2.59 & 4.93 & 7.96 & 16.21 & 8.37 & 16.15 \\

    w/o Synergy &
    2.45 & 4.41 & 2.55 & 4.81 & 7.72 & 15.95 & 8.39 & 16.19 \\

    Shared Weight &
    2.46 & 4.50 & 2.59 & 4.82 & 8.07 & 16.62 & 8.57 & 16.62 \\    

    w/o Node Emb &
    2.52 & 4.58 & 2.59 & 4.87 & 8.44 & 17.08 & 8.77 & 17.09 \\

    \textbf{FreMo} &
    \textbf{2.42} & \textbf{4.39} & \textbf{2.55} & \textbf{4.80} & \textbf{7.54} & \textbf{15.78} & \textbf{8.25} & \textbf{15.86} \\
    
    \bottomrule
    \end{tabular}%
}
\label{tab:ablation}
\end{table}

\subsection{Ablation Study}
\subsubsection{Effect of Each Component.}
To evaluate the contribution of each component in FreMo, ablation studies are conducted on the NYC dataset (Table~\ref{tab:ablation}) with the following variants:

(1) \emph{w/o FreMo}: removes the entire FreMo module and uses the spatio-temporal encoder as the backbone.

(2) \emph{Time Domain}: replaces the frequency-domain operations in FreMo with equivalent time-domain processing blocks.

(3) \emph{w/o Synergy}: removes the FSI and keeps only the modality-wise refinement (MFF).

(4) \emph{Shared Weight}: replaces the modality-wise weight generators with a single shared generator across modalities.

(5) \emph{w/o Node Emb}: removes the node embedding used in MFF.
\begin{figure}[t]
  \centering
  \includegraphics[width=\linewidth]{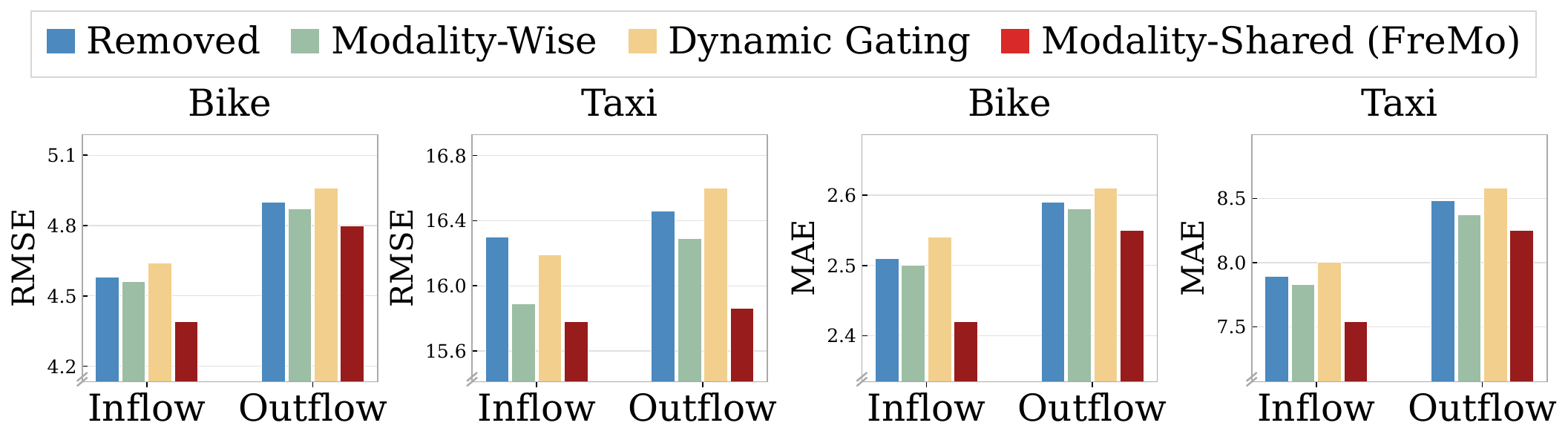}
  \caption{Performance of different synergy injection strategies (parameterization of $\gamma$) on the NYC.}
  \Description{Bar chart of ablation studies on the impact of the synergy scalar.}
  \label{figure:ablation_gamma}
\end{figure}
Table~\ref{tab:ablation} yields the following observations:
(1) Removing FreMo (\emph{w/o FreMo}) degrades all modalities, indicating that the frequency-domain refinement and synergy modeling in FreMo provide substantial gains beyond the encoder alone.
(2) \emph{Time Domain} underperforms FreMo, suggesting that explicitly operating in the frequency domain better captures scale-dependent structures than purely time-domain modeling.
(3) Removing FSI (\emph{w/o Synergy}) causes minor changes on Bike flows but larger drops on Taxi flows, implying that frequency-wise synergy consensus is particularly helpful for modalities that benefit more from cross-modality consistent components.
(4) \emph{Shared Weight} consistently degrades performance (especially on Taxi), showing that using separate modality-wise generators is important to accommodate modality-dependent spectral profiles, rather than enforcing a single shared parameterization.
(5) \emph{w/o Node Emb} performs worse, highlighting that incorporating node embeddings helps MFF adapt its frequency gates to different spatial units, which improves the effectiveness of frequency refinement.
Overall, the gains of FreMo come from the combined contributions of MFF and FSI, rather than any single design in isolation.

\subsubsection{Impact of Synergy Scalar $\gamma$.}
Figure~\ref{figure:ablation_gamma} examines how the residual scaling factor $\gamma$ of the synergy consensus affects the feedback injection into modality-wise representations. \emph{Removed} sets $\gamma=1$ (i.e., injecting $\mathcal{C}$ without learnable scaling), \emph{Modality-Wise} learns an independent scalar for each modality, \emph{Dynamic Gating} uses a two-layer MLP to generate the gating coefficient, and \emph{Modality-Shared (FreMo)} uses a single learnable scalar shared across modalities.
(\emph{Removed}) consistently degrades performance, indicating that a learnable coefficient is important for balancing the original features and the injected consensus. 
Both \emph{Modality-Wise} and \emph{Dynamic Gating} lead to worse results across metrics, and \emph{Dynamic Gating} is the most unstable choice. This suggests that introducing extra flexibility in the feedback (either per-modality parameters or dynamically generated gates) brings limited benefit and can make optimization harder.
In contrast, (\emph{Modality-Shared}) achieves the best performance, providing a simple and effective global control of synergy injection with minimal parameter overhead.

\begin{table}[t]
\centering
\caption{Applying FreMo to representative time series models on the NYC. Results are taken from the third horizon.}
\setlength{\tabcolsep}{1pt}
\resizebox{\linewidth}{!}{%
    \begin{tabular}{c|cc|cc|cc|cc}
    \toprule    
    \multirow{2}{*}{Model} & 
    \multicolumn{2}{c|}{Bike Inflow} & 
    \multicolumn{2}{c|}{Bike Outflow} & 
    \multicolumn{2}{c|}{Taxi Inflow} & 
    \multicolumn{2}{c}{Taxi Outflow} \\
    
    \cmidrule(lr){2-3} \cmidrule(lr){4-5} \cmidrule(lr){6-7} \cmidrule(lr){8-9}

     & MAE & RMSE & MAE & RMSE & MAE & RMSE & MAE & RMSE \\
    \midrule
    
    AGCRN~\cite{bai2020adaptive} &
    3.02 & 8.41 & 3.31 & 9.15 & 9.16 & 18.16 & 10.19 & 19.29 \\
    \textbf{+FreMo} &
    \textbf{2.89} & \textbf{5.78} & \textbf{3.27} & \textbf{6.18} & \textbf{8.84} & \textbf{16.69} & \textbf{9.75} & \textbf{18.07} \\
    $\Delta$(\%) &
    \textbf{4.30\%} & \textbf{31.27\%} & \textbf{1.21\%} & \textbf{32.46\%} & \textbf{3.49\%} & \textbf{8.09\%} & \textbf{4.32\%} & \textbf{6.32\%} \\
    \midrule

    TimesNet~\cite{wu2023timesnet}&
    3.21 & 5.88 & 3.45 & 6.34 & 9.25 & 18.04 & 10.57 & 20.37 \\
    \textbf{+FreMo} &
    \textbf{3.15} & \textbf{5.78} & \textbf{3.41} & \textbf{6.26} & \textbf{9.22} & \textbf{17.95} & \textbf{10.35} & \textbf{20.02} \\
    $\Delta$(\%) &
    \textbf{1.87\%} & \textbf{1.70\%} & \textbf{1.16\%} & \textbf{1.26\%} & \textbf{0.32\%} & \textbf{0.50\%} & \textbf{2.08\%} & \textbf{1.72\%} \\
    \midrule
    
    iTransformer~\cite{liu2024itransformer}&
    4.52 & 8.85 & 4.80 & 9.44 & 12.54 & 23.98 & 12.80 & 24.37 \\
    \textbf{+FreMo} &
    \textbf{3.77} & \textbf{6.88} & \textbf{3.94} & \textbf{7.58} & \textbf{10.66} & \textbf{19.81} & \textbf{11.38} & \textbf{20.75} \\
    $\Delta$(\%) &
    \textbf{16.59\%} & \textbf{22.26\%} & \textbf{17.92\%} & \textbf{19.70\%} & \textbf{14.99\%} & \textbf{17.39\%} & \textbf{11.09\%} & \textbf{14.85\%} \\
    \midrule

    STAEformer~\cite{liu2023spatio}& 
    3.81 & 7.78 & 3.73 & 7.74 & 11.89 & 24.06 & 12.14 & 21.75 \\
    \textbf{+FreMo} &
    \textbf{3.10} & \textbf{6.32} & \textbf{3.24} & \textbf{7.07} & \textbf{9.26} & \textbf{19.02} & \textbf{9.46} & \textbf{18.48} \\
    $\Delta$(\%) &
    \textbf{18.64\%} & \textbf{18.77\%} & \textbf{13.14\%} & \textbf{8.66\%} & \textbf{22.12\%} & \textbf{20.95\%} & \textbf{22.08\%} & \textbf{15.03\%} \\
    
    \bottomrule
    \end{tabular}%
}
\label{tab:generality}
\end{table}

\subsubsection{Plug-and-Play Capacity}
To evaluate the plug-and-play capability, we integrate FreMo into four representative backbones, including AGCRN, TimesNet, iTransformer, and STAEformer. FreMo is inserted as an auxiliary enhancement module without modifying the original backbone architecture. Full results are provided in Appendix~\ref{appendix_generality}.
Table~\ref{tab:generality} shows that FreMo consistently improves performance across all backbones, modalities, and metrics. We make the following observations:
(1) FreMo brings substantial gains on Transformer-based backbones. For iTransformer, FreMo achieves double-digit improvements, reducing MAE by $11.09\%-17.92\%$ and RMSE by $14.85\%-22.26\%$. For STAEformer, the gains are more pronounced on Taxi flows, with about $22.1\%$ MAE reduction and $15.03\%-20.95\%$ RMSE reduction.
(2) FreMo improves AGCRN notably in RMSE. MAE reductions are modest (around $1.21\%-4.30\%$), while RMSE on Bike flows drops by $31.27\%-32.46\%$, suggesting fewer large-error cases.
(3) FreMo remains effective with a frequency-based backbone. Although TimesNet incorporates frequency-related modeling, FreMo still provides consistent gains, typically around $1\%-2\%$ on both MAE and RMSE, indicating complementary benefits.
These results are consistent with the motivation of FreMo: common backbones often treat modalities as independent channels or simply concatenated inputs, while FreMo explicitly refines frequency components for each modality and enables selective cross-modality synergy. Overall, FreMo serves as an architecture-agnostic enhancer that can be plugged into diverse backbones to improve multi-modality forecasting.

\begin{figure}[t]
  \centering
  \includegraphics[width=\linewidth]{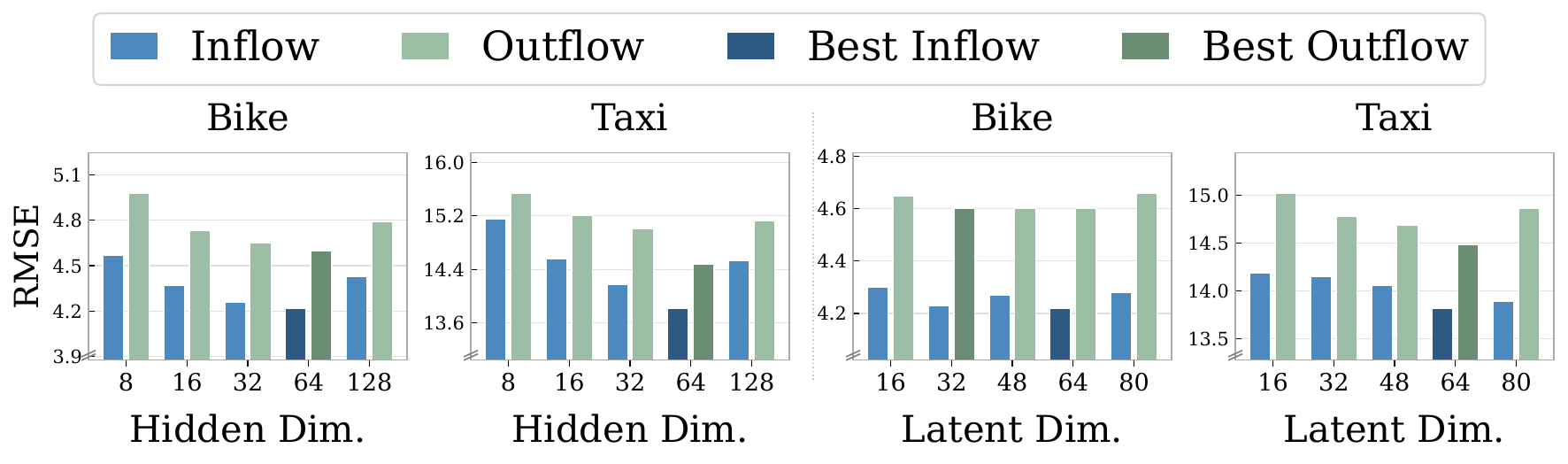}
  \caption{Hyperparameter sensitivity on the NYC.}
  \Description{Bar chart of hyperparameter studies on the NYC dataset.}
  \label{figure:hyper_parameter_nyc}
\end{figure}

\subsubsection{Hyperparameter Sensitivity}
As shown in Figure~\ref{figure:hyper_parameter_nyc}, the performance of FreMo on the NYC dataset is influenced by two considered hyperparameters, i.e., the hidden dimension $d$ and the latent dimension $L$. As hidden dimension increases, the performance first decreases and then rises, consistent with the underfitting-overfitting trade-off. The best result is achieved at $d=64$. For the latent dimension, Taxi flows exhibit a similar pattern with the optimum at $L=64$, whereas Bike flows are relatively stable across $L \in \{32,48,64\}$. Complete results on all datasets are provided in Appendix~\ref{appendix_hyperparameter}.

\begin{figure}[t]
  \centering
  \includegraphics[width=\linewidth]{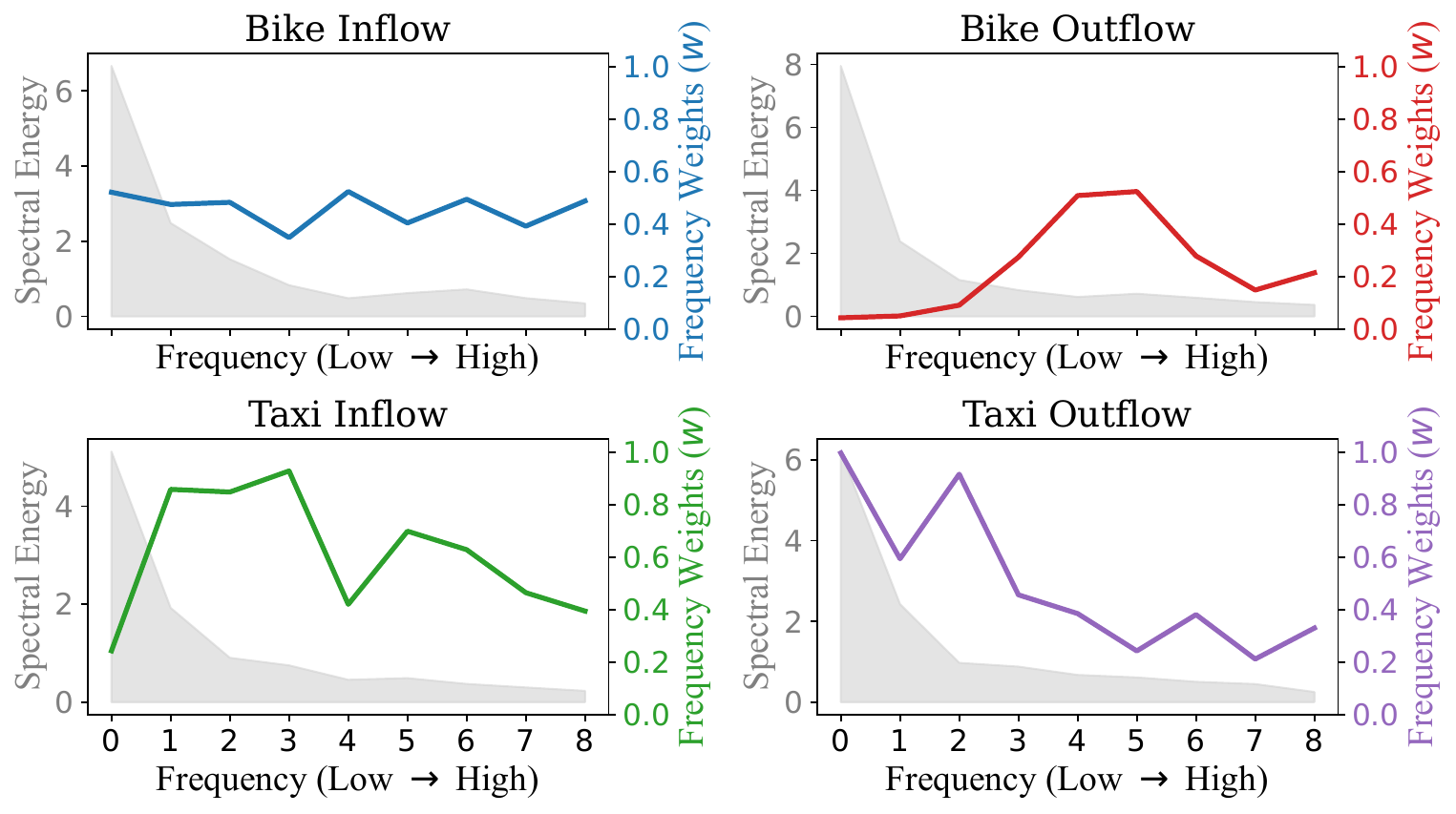}
  \caption{The learned modality-wise frequency weights on the NYC. Gray shaded areas denote the raw spectral amplitude $A$ (left axis), and the colored curves show the frequency weights $w$ (right axis).}
  \Description{Line chart of the learned modality-wise frequency weights.}
  \label{figure:case01_weights}
\end{figure}
\subsection{Case Study}
\subsubsection{Modality-wise Frequency Weights.}
To examine whether MFF can identify informative frequency components within each modality, we visualize the learned frequency weights $w$ in Eq.~(\ref{eq:fre_weight}) on a representative node, as shown in Figure~\ref{figure:case01_weights}. The gray shaded area denotes the spectral amplitude, while the colored curve shows the frequency weights (right axis), indicating how strongly each frequency bin is retained after filtering. 
Across modalities, the raw spectral amplitude exhibits a similar low-to-high decay trend, reflecting the low-frequency dominance of traffic signals. However, the learned frequency responses differ clearly across modalities. Bike Inflow shows a relatively balanced response across frequencies, whereas Bike Outflow places more emphasis on mid-frequency components. Taxi Inflow exhibits strong low-frequency responses with an additional mid-frequency emphasis, followed by attenuation at high frequencies. Taxi Outflow is highly concentrated on the lowest frequencies and decays rapidly thereafter.
These results suggest that spectral amplitude alone is not sufficient to determine frequency importance. If filtering were solely driven by amplitude magnitude, the model would tend to overemphasize low-frequency components and under-utilize informative mid/high-frequency patterns. Instead, even under a similar amplitude decay trend, FreMo learns distinct modality-wise frequency gates, supporting the motivation that different modalities benefit from different frequency-band selections.

\begin{figure}[t]
  \centering
  \includegraphics[width=\linewidth]{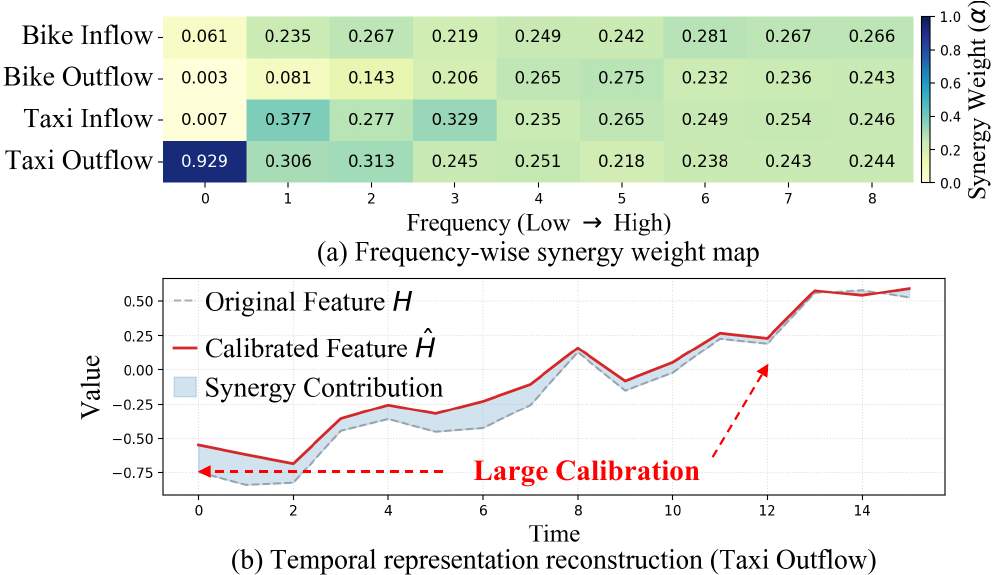}
  \caption{The frequency-wise synergy weight and selective temporal calibration on the NYC. (a) Synergy weight $\alpha$ indicates each modality’s relative contribution at each frequency bin. (b) Temporal comparison between original ($H$) and synergy-calibrated ($\hat{H}$) Taxi Outflow representations.}
  \Description{(a) A frequency-bin and modality heatmap of synergy weights. (b) Time-domain comparison of original and synergy-calibrated representations for Taxi Outflow.}
  \label{figure:case02_synergy}
\end{figure}

\subsubsection{Frequency-wise Synergy Weights and Selective Temporal Calibration.}
This case study examines how FSI assigns frequency-wise synergy weights across modalities and how the resulting feedback leads to selective temporal calibration. We visualize the Softmax-normalized synergy weights $\alpha$ in Eq.~(\ref{eq:score}) over frequency bins, and then compare the original temporal feature $H$ with the synergy-calibrated feature $\hat{H}$ for Taxi Outflow.

Figure~\ref{figure:case02_synergy}(a) shows the frequency-wise synergy weight map on a representative node, where each entry $\alpha$ indicates the relative contribution of a modality at a specific frequency bin. At the lowest-frequency bin, the weights are highly concentrated on Taxi Outflow, suggesting that the synergy consensus in this regime is mainly driven by Taxi-related signals. As frequency increases, the weights become less concentrated and spread more evenly across modalities, indicating that no single modality consistently dominates in higher-frequency components.
Figure~\ref{figure:case02_synergy}(b) further illustrates the temporal effect of synergy feedback on Taxi Outflow. Synergy feedback does not alter the sequence uniformly. Larger calibrations appear in the earlier segment, while the later segment shows much smaller calibrations and the two curves stay close. This indicates that FreMo performs selective temporal calibration rather than applying a global smoothing effect, strengthening the representation only when it is needed. Overall, FreMo learns frequency-wise synergy weights and translates them into interpretable, adaptive refinement in the temporal domain.

\begin{figure}[t]
  \centering
  \includegraphics[width=\linewidth]{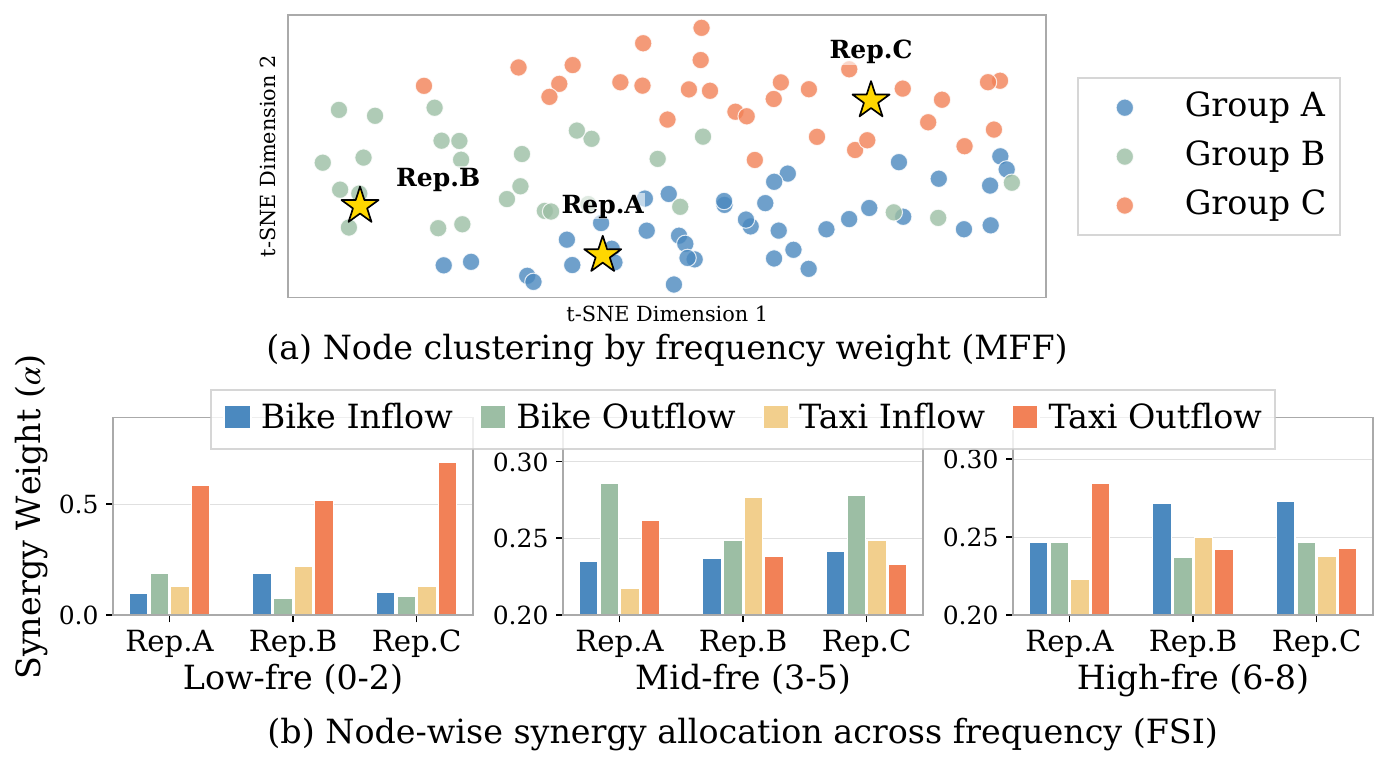}
  \caption{Node-level frequency gating and synergy allocation. (a) Node clustering by frequency weight $w$. (b) Synergy weights $\alpha$ of representative nodes across frequencies.}
  \Description{visualization of node clustering by frequency weight via t-SNE and synergy weights of representative nodes across frequencies via bar chat.}
  \label{figure:case03_node}
\end{figure}

\subsubsection{Node-level Frequency Gating and Synergy Allocation.}
To examine whether FreMo exhibits structured node-wise differences in frequency filtering, we cluster nodes using the modality-wise frequency weight $w$ and visualize them with t-SNE, as shown in Figure~\ref{figure:case03_node}(a). The separable clustering trend suggests that MFF learns node-dependent frequency-gating profiles rather than applying a uniform gate. 
We further inspect frequency-wise synergy weights ($\alpha$) on representative nodes (Rep.A/Rep.B/Rep.C) selected from different clusters, as shown in Figure~\ref{figure:case03_node}(b). The nodes exhibit frequency-dependent variations in modality allocation, especially in the mid/high-frequency bands.
This highlights that FreMo learns node-dependent frequency gating, and the resulting node diversity manifests as frequency-dependent variations in synergy allocation.

\section{Conclusion}
In this paper, we propose a frequency-domain modeling method (FreMo) for multi-modality transportation forecasting, designed to achieve outstanding performance with a simple architecture. FreMo decouples the modeling pipeline into two complementary paradigms by leveraging frequency-wise discriminative properties: a Modality-Wise Frequency Filter that accentuates informative spectral patterns within each modality without compromising temporal alignment, and a Frequency-Guided Synergy Integrator that coordinates cross-modal synergy by dynamically weighting reliable modality contributions at each frequency. Experiments on several multi-modality transportation datasets demonstrate that FreMo achieves state-of-the-art forecasting performance with robust generalization across diverse scenarios.

\bibliographystyle{ACM-Reference-Format}
\bibliography{reference}


\begin{thebibliography}{80}


\ifx \showCODEN    \undefined \def \showCODEN     #1{\unskip}     \fi
\ifx \showISBNx    \undefined \def \showISBNx     #1{\unskip}     \fi
\ifx \showISBNxiii \undefined \def \showISBNxiii  #1{\unskip}     \fi
\ifx \showISSN     \undefined \def \showISSN      #1{\unskip}     \fi
\ifx \showLCCN     \undefined \def \showLCCN      #1{\unskip}     \fi
\ifx \shownote     \undefined \def \shownote      #1{#1}          \fi
\ifx \showarticletitle \undefined \def \showarticletitle #1{#1}   \fi
\ifx \showURL      \undefined \def \showURL       {\relax}        \fi
\providecommand\bibfield[2]{#2}
\providecommand\bibinfo[2]{#2}
\providecommand\natexlab[1]{#1}
\providecommand\showeprint[2][]{arXiv:#2}

\bibitem[Ba et~al\mbox{.}(2016)]%
        {ba2016layer}
\bibfield{author}{\bibinfo{person}{Jimmy~Lei Ba}, \bibinfo{person}{Jamie~Ryan Kiros}, {and} \bibinfo{person}{Geoffrey~E Hinton}.} \bibinfo{year}{2016}\natexlab{}.
\newblock \showarticletitle{Layer normalization}.
\newblock \bibinfo{journal}{\emph{arXiv preprint arXiv:1607.06450}} (\bibinfo{year}{2016}).
\newblock


\bibitem[Bai et~al\mbox{.}(2020)]%
        {bai2020adaptive}
\bibfield{author}{\bibinfo{person}{Lei Bai}, \bibinfo{person}{Lina Yao}, \bibinfo{person}{Can Li}, \bibinfo{person}{Xianzhi Wang}, {and} \bibinfo{person}{Can Wang}.} \bibinfo{year}{2020}\natexlab{}.
\newblock \showarticletitle{Adaptive graph convolutional recurrent network for traffic forecasting}.
\newblock \bibinfo{journal}{\emph{Advances in Neural Information Processing Systems}}  \bibinfo{volume}{33} (\bibinfo{year}{2020}), \bibinfo{pages}{17804--17815}.
\newblock


\bibitem[Cai et~al\mbox{.}(2024)]%
        {cai2024msgnet}
\bibfield{author}{\bibinfo{person}{Wanlin Cai}, \bibinfo{person}{Yuxuan Liang}, \bibinfo{person}{Xianggen Liu}, \bibinfo{person}{Jianshuai Feng}, {and} \bibinfo{person}{Yuankai Wu}.} \bibinfo{year}{2024}\natexlab{}.
\newblock \showarticletitle{Msgnet: Learning multi-scale inter-series correlations for multivariate time series forecasting}. In \bibinfo{booktitle}{\emph{Proceedings of the AAAI Conference on Artificial Intelligence}}, Vol.~\bibinfo{volume}{38}. \bibinfo{pages}{11141--11149}.
\newblock


\bibitem[Chen et~al\mbox{.}(2024)]%
        {chen2024pathformer}
\bibfield{author}{\bibinfo{person}{Peng Chen}, \bibinfo{person}{Yingying Zhang}, \bibinfo{person}{Yunyao Cheng}, \bibinfo{person}{Yang Shu}, \bibinfo{person}{Yihang Wang}, \bibinfo{person}{Qingsong Wen}, \bibinfo{person}{Bin Yang}, {and} \bibinfo{person}{Chenjuan Guo}.} \bibinfo{year}{2024}\natexlab{}.
\newblock \showarticletitle{Pathformer: Multi-scale Transformers with Adaptive Pathways for Time Series Forecasting}. In \bibinfo{booktitle}{\emph{International Conference on Learning Representations}}.
\newblock


\bibitem[Cheng et~al\mbox{.}(2023)]%
        {cheng2023formertime}
\bibfield{author}{\bibinfo{person}{Mingyue Cheng}, \bibinfo{person}{Qi Liu}, \bibinfo{person}{Zhiding Liu}, \bibinfo{person}{Zhi Li}, \bibinfo{person}{Yucong Luo}, {and} \bibinfo{person}{Enhong Chen}.} \bibinfo{year}{2023}\natexlab{}.
\newblock \showarticletitle{Formertime: Hierarchical multi-scale representations for multivariate time series classification}. In \bibinfo{booktitle}{\emph{Proceedings of the ACM web conference}}. \bibinfo{pages}{1437--1445}.
\newblock


\bibitem[Dai et~al\mbox{.}(2024)]%
        {dai2024periodicity}
\bibfield{author}{\bibinfo{person}{Tao Dai}, \bibinfo{person}{Beiliang Wu}, \bibinfo{person}{Peiyuan Liu}, \bibinfo{person}{Naiqi Li}, \bibinfo{person}{Jigang Bao}, \bibinfo{person}{Yong Jiang}, {and} \bibinfo{person}{Shu-Tao Xia}.} \bibinfo{year}{2024}\natexlab{}.
\newblock \showarticletitle{Periodicity decoupling framework for long-term series forecasting}. In \bibinfo{booktitle}{\emph{International Conference on Learning Representations}}.
\newblock


\bibitem[Deldari et~al\mbox{.}(2022)]%
        {deldari2022cocoa}
\bibfield{author}{\bibinfo{person}{Shohreh Deldari}, \bibinfo{person}{Hao Xue}, \bibinfo{person}{Aaqib Saeed}, \bibinfo{person}{Daniel~V Smith}, {and} \bibinfo{person}{Flora~D Salim}.} \bibinfo{year}{2022}\natexlab{}.
\newblock \showarticletitle{Cocoa: Cross modality contrastive learning for sensor data}.
\newblock \bibinfo{journal}{\emph{Proceedings of the ACM on Interactive, Mobile, Wearable and Ubiquitous Technologies}} \bibinfo{volume}{6}, \bibinfo{number}{3} (\bibinfo{year}{2022}), \bibinfo{pages}{1--28}.
\newblock


\bibitem[Deng et~al\mbox{.}(2023a)]%
        {deng2023gmrl}
\bibfield{author}{\bibinfo{person}{Jiewen Deng}, \bibinfo{person}{Jinliang Deng}, \bibinfo{person}{Renhe Jiang}, {and} \bibinfo{person}{Xuan Song}.} \bibinfo{year}{2023}\natexlab{a}.
\newblock \showarticletitle{Learning Gaussian Mixture Representations for Tensor Time Series Forecasting}. In \bibinfo{booktitle}{\emph{Proceedings of the International Joint Conference on Artificial Intelligence}}. \bibinfo{pages}{2077--2085}.
\newblock


\bibitem[Deng et~al\mbox{.}(2023b)]%
        {deng2023tts}
\bibfield{author}{\bibinfo{person}{Jiewen Deng}, \bibinfo{person}{Jinliang Deng}, \bibinfo{person}{Du Yin}, \bibinfo{person}{Renhe Jiang}, {and} \bibinfo{person}{Xuan Song}.} \bibinfo{year}{2023}\natexlab{b}.
\newblock \showarticletitle{Tts-norm: Forecasting tensor time series via multi-way normalization}.
\newblock \bibinfo{journal}{\emph{ACM Transactions on Knowledge Discovery from Data}} \bibinfo{volume}{18}, \bibinfo{number}{1} (\bibinfo{year}{2023}), \bibinfo{pages}{1--25}.
\newblock


\bibitem[Deng et~al\mbox{.}(2024)]%
        {deng2024mossl}
\bibfield{author}{\bibinfo{person}{Jiewen Deng}, \bibinfo{person}{Renhe Jiang}, \bibinfo{person}{Jiaqi Zhang}, {and} \bibinfo{person}{Xuan Song}.} \bibinfo{year}{2024}\natexlab{}.
\newblock \showarticletitle{Multi-Modality Spatio-Temporal Forecasting via Self-Supervised Learning}. In \bibinfo{booktitle}{\emph{Proceedings of the International Joint Conference on Artificial Intelligence}}. \bibinfo{pages}{2018--2026}.
\newblock


\bibitem[Deng et~al\mbox{.}(2022)]%
        {deng2022graph}
\bibfield{author}{\bibinfo{person}{Leyan Deng}, \bibinfo{person}{Defu Lian}, \bibinfo{person}{Zhenya Huang}, {and} \bibinfo{person}{Enhong Chen}.} \bibinfo{year}{2022}\natexlab{}.
\newblock \showarticletitle{Graph convolutional adversarial networks for spatiotemporal anomaly detection}.
\newblock \bibinfo{journal}{\emph{IEEE Transactions on Neural Networks and Learning Systems}} \bibinfo{volume}{33}, \bibinfo{number}{6} (\bibinfo{year}{2022}), \bibinfo{pages}{2416--2428}.
\newblock


\bibitem[Fan et~al\mbox{.}(2023)]%
        {fan2023dish}
\bibfield{author}{\bibinfo{person}{Wei Fan}, \bibinfo{person}{Pengyang Wang}, \bibinfo{person}{Dongkun Wang}, \bibinfo{person}{Dongjie Wang}, \bibinfo{person}{Yuanchun Zhou}, {and} \bibinfo{person}{Yanjie Fu}.} \bibinfo{year}{2023}\natexlab{}.
\newblock \showarticletitle{Dish-ts: a general paradigm for alleviating distribution shift in time series forecasting}. In \bibinfo{booktitle}{\emph{Proceedings of the AAAI Conference on Artificial Intelligence}}, Vol.~\bibinfo{volume}{37}. \bibinfo{pages}{7522--7529}.
\newblock


\bibitem[Fang et~al\mbox{.}(2019)]%
        {fang2019gstnet}
\bibfield{author}{\bibinfo{person}{Shen Fang}, \bibinfo{person}{Qi Zhang}, \bibinfo{person}{Gaofeng Meng}, \bibinfo{person}{Shiming Xiang}, {and} \bibinfo{person}{Chunhong Pan}.} \bibinfo{year}{2019}\natexlab{}.
\newblock \showarticletitle{GSTNet: Global Spatial-Temporal Network for Traffic Flow Prediction.}. In \bibinfo{booktitle}{\emph{IJCAI}}. \bibinfo{pages}{2286--2293}.
\newblock


\bibitem[Fang et~al\mbox{.}(2025)]%
        {fang2025efficient}
\bibfield{author}{\bibinfo{person}{Yuchen Fang}, \bibinfo{person}{Yuxuan Liang}, \bibinfo{person}{Bo Hui}, \bibinfo{person}{Zezhi Shao}, \bibinfo{person}{Liwei Deng}, \bibinfo{person}{Xu Liu}, \bibinfo{person}{Xinke Jiang}, {and} \bibinfo{person}{Kai Zheng}.} \bibinfo{year}{2025}\natexlab{}.
\newblock \showarticletitle{Efficient large-scale traffic forecasting with transformers: A spatial data management perspective}. In \bibinfo{booktitle}{\emph{Proceedings of the ACM SIGKDD Conference on Knowledge Discovery and Data Mining}}. \bibinfo{pages}{307--317}.
\newblock


\bibitem[Fang et~al\mbox{.}(2021)]%
        {fang2021mdtp}
\bibfield{author}{\bibinfo{person}{Ziquan Fang}, \bibinfo{person}{Lu Pan}, \bibinfo{person}{Lu Chen}, \bibinfo{person}{Yuntao Du}, {and} \bibinfo{person}{Yunjun Gao}.} \bibinfo{year}{2021}\natexlab{}.
\newblock \showarticletitle{MDTP: a multi-source deep traffic prediction framework over spatio-temporal trajectory data}.
\newblock \bibinfo{journal}{\emph{Proceedings of the VLDB Endowment}} \bibinfo{volume}{14}, \bibinfo{number}{8} (\bibinfo{year}{2021}), \bibinfo{pages}{1289--1297}.
\newblock


\bibitem[Fu and Hu(2025)]%
        {fu2025frequency}
\bibfield{author}{\bibinfo{person}{En Fu} {and} \bibinfo{person}{Yanyan Hu}.} \bibinfo{year}{2025}\natexlab{}.
\newblock \showarticletitle{Frequency-Masked Embedding Inference: A Non-Contrastive Approach for Time Series Representation Learning}. In \bibinfo{booktitle}{\emph{Proceedings of the AAAI Conference on Artificial Intelligence}}, Vol.~\bibinfo{volume}{39}. \bibinfo{pages}{16639--16647}.
\newblock


\bibitem[Guo et~al\mbox{.}(2021)]%
        {guo2021hierarchical}
\bibfield{author}{\bibinfo{person}{Kan Guo}, \bibinfo{person}{Yongli Hu}, \bibinfo{person}{Yanfeng Sun}, \bibinfo{person}{Sean Qian}, \bibinfo{person}{Junbin Gao}, {and} \bibinfo{person}{Baocai Yin}.} \bibinfo{year}{2021}\natexlab{}.
\newblock \showarticletitle{Hierarchical Graph Convolution Network for Traffic Forecasting}. In \bibinfo{booktitle}{\emph{Proceedings of the AAAI Conference on Artificial Intelligence}}, Vol.~\bibinfo{volume}{35}. \bibinfo{pages}{151--159}.
\newblock


\bibitem[Guo et~al\mbox{.}(2019a)]%
        {guo2019attention}
\bibfield{author}{\bibinfo{person}{Shengnan Guo}, \bibinfo{person}{Youfang Lin}, \bibinfo{person}{Ning Feng}, \bibinfo{person}{Chao Song}, {and} \bibinfo{person}{Huaiyu Wan}.} \bibinfo{year}{2019}\natexlab{a}.
\newblock \showarticletitle{Attention based spatial-temporal graph convolutional networks for traffic flow forecasting}. In \bibinfo{booktitle}{\emph{Proceedings of the AAAI Conference on Artificial Intelligence}}, Vol.~\bibinfo{volume}{33}. \bibinfo{pages}{922--929}.
\newblock


\bibitem[Guo et~al\mbox{.}(2019b)]%
        {guo2019deep}
\bibfield{author}{\bibinfo{person}{Shengnan Guo}, \bibinfo{person}{Youfang Lin}, \bibinfo{person}{Shijie Li}, \bibinfo{person}{Zhaoming Chen}, {and} \bibinfo{person}{Huaiyu Wan}.} \bibinfo{year}{2019}\natexlab{b}.
\newblock \showarticletitle{Deep spatial--temporal 3D convolutional neural networks for traffic data forecasting}.
\newblock \bibinfo{journal}{\emph{IEEE Transactions on Intelligent Transportation Systems}} \bibinfo{volume}{20}, \bibinfo{number}{10} (\bibinfo{year}{2019}), \bibinfo{pages}{3913--3926}.
\newblock


\bibitem[Han et~al\mbox{.}(2021b)]%
        {han2021joint}
\bibfield{author}{\bibinfo{person}{Jindong Han}, \bibinfo{person}{Hao Liu}, \bibinfo{person}{Hengshu Zhu}, \bibinfo{person}{Hui Xiong}, {and} \bibinfo{person}{Dejing Dou}.} \bibinfo{year}{2021}\natexlab{b}.
\newblock \showarticletitle{Joint air quality and weather prediction based on multi-adversarial spatiotemporal networks}. In \bibinfo{booktitle}{\emph{Proceedings of the AAAI Conference on Artificial Intelligence}}, Vol.~\bibinfo{volume}{35}. \bibinfo{pages}{4081--4089}.
\newblock


\bibitem[Han et~al\mbox{.}(2021a)]%
        {han2021dynamic}
\bibfield{author}{\bibinfo{person}{Liangzhe Han}, \bibinfo{person}{Bowen Du}, \bibinfo{person}{Leilei Sun}, \bibinfo{person}{Yanjie Fu}, \bibinfo{person}{Yisheng Lv}, {and} \bibinfo{person}{Hui Xiong}.} \bibinfo{year}{2021}\natexlab{a}.
\newblock \showarticletitle{Dynamic and multi-faceted spatio-temporal deep learning for traffic speed forecasting}. In \bibinfo{booktitle}{\emph{Proceedings of the ACM SIGKDD Conference on Knowledge Discovery and Data Mining}}. \bibinfo{pages}{547--555}.
\newblock


\bibitem[Huang et~al\mbox{.}(2019)]%
        {huang2019mist}
\bibfield{author}{\bibinfo{person}{Chao Huang}, \bibinfo{person}{Chuxu Zhang}, \bibinfo{person}{Jiashu Zhao}, \bibinfo{person}{Xian Wu}, \bibinfo{person}{Dawei Yin}, {and} \bibinfo{person}{Nitesh Chawla}.} \bibinfo{year}{2019}\natexlab{}.
\newblock \showarticletitle{Mist: A multiview and multimodal spatial-temporal learning framework for citywide abnormal event forecasting}. In \bibinfo{booktitle}{\emph{The World Wide Web conference}}. \bibinfo{pages}{717--728}.
\newblock


\bibitem[Huang et~al\mbox{.}(2018)]%
        {huang2018deepcrime}
\bibfield{author}{\bibinfo{person}{Chao Huang}, \bibinfo{person}{Junbo Zhang}, \bibinfo{person}{Yu Zheng}, {and} \bibinfo{person}{Nitesh~V. Chawla}.} \bibinfo{year}{2018}\natexlab{}.
\newblock \showarticletitle{DeepCrime: Attentive Hierarchical Recurrent Networks for Crime Prediction}. In \bibinfo{booktitle}{\emph{Proceedings of the ACM International Conference on Information and Knowledge Management}}. \bibinfo{pages}{1423--1432}.
\newblock


\bibitem[Huang et~al\mbox{.}(2024)]%
        {huang2024hdmixer}
\bibfield{author}{\bibinfo{person}{Qihe Huang}, \bibinfo{person}{Lei Shen}, \bibinfo{person}{Ruixin Zhang}, \bibinfo{person}{Jiahuan Cheng}, \bibinfo{person}{Shouhong Ding}, \bibinfo{person}{Zhengyang Zhou}, {and} \bibinfo{person}{Yang Wang}.} \bibinfo{year}{2024}\natexlab{}.
\newblock \showarticletitle{Hdmixer: Hierarchical dependency with extendable patch for multivariate time series forecasting}. In \bibinfo{booktitle}{\emph{Proceedings of the AAAI Conference on Artificial Intelligence}}, Vol.~\bibinfo{volume}{38}. \bibinfo{pages}{12608--12616}.
\newblock


\bibitem[Huang et~al\mbox{.}(2023)]%
        {huang2023crossgnn}
\bibfield{author}{\bibinfo{person}{Qihe Huang}, \bibinfo{person}{Lei Shen}, \bibinfo{person}{Ruixin Zhang}, \bibinfo{person}{Shouhong Ding}, \bibinfo{person}{Binwu Wang}, \bibinfo{person}{Zhengyang Zhou}, {and} \bibinfo{person}{Yang Wang}.} \bibinfo{year}{2023}\natexlab{}.
\newblock \showarticletitle{Crossgnn: Confronting noisy multivariate time series via cross interaction refinement}.
\newblock \bibinfo{journal}{\emph{Advances in Neural Information Processing Systems}}  \bibinfo{volume}{36} (\bibinfo{year}{2023}), \bibinfo{pages}{46885--46902}.
\newblock


\bibitem[Ji et~al\mbox{.}(2023)]%
        {ji2023spatio}
\bibfield{author}{\bibinfo{person}{Jiahao Ji}, \bibinfo{person}{Jingyuan Wang}, \bibinfo{person}{Chao Huang}, \bibinfo{person}{Junjie Wu}, \bibinfo{person}{Boren Xu}, \bibinfo{person}{Zhenhe Wu}, \bibinfo{person}{Junbo Zhang}, {and} \bibinfo{person}{Yu Zheng}.} \bibinfo{year}{2023}\natexlab{}.
\newblock \showarticletitle{Spatio-temporal self-supervised learning for traffic flow prediction}. In \bibinfo{booktitle}{\emph{Proceedings of the AAAI Conference on Artificial Intelligence}}, Vol.~\bibinfo{volume}{37}. \bibinfo{pages}{4356--4364}.
\newblock


\bibitem[Jiang et~al\mbox{.}(2023)]%
        {jiang2023pdformer}
\bibfield{author}{\bibinfo{person}{Jiawei Jiang}, \bibinfo{person}{Chengkai Han}, \bibinfo{person}{Wayne~Xin Zhao}, {and} \bibinfo{person}{Jingyuan Wang}.} \bibinfo{year}{2023}\natexlab{}.
\newblock \showarticletitle{Pdformer: Propagation delay-aware dynamic long-range transformer for traffic flow prediction}. In \bibinfo{booktitle}{\emph{Proceedings of the AAAI Conference on Artificial Intelligence}}, Vol.~\bibinfo{volume}{37}. \bibinfo{pages}{4365--4373}.
\newblock


\bibitem[Jiang et~al\mbox{.}(2021a)]%
        {jiang2021deepcrowd}
\bibfield{author}{\bibinfo{person}{Renhe Jiang}, \bibinfo{person}{Zekun Cai}, \bibinfo{person}{Zhaonan Wang}, \bibinfo{person}{Chuang Yang}, \bibinfo{person}{Zipei Fan}, \bibinfo{person}{Quanjun Chen}, \bibinfo{person}{Kota Tsubouchi}, \bibinfo{person}{Xuan Song}, {and} \bibinfo{person}{Ryosuke Shibasaki}.} \bibinfo{year}{2021}\natexlab{a}.
\newblock \showarticletitle{DeepCrowd: A deep model for large-scale citywide crowd density and flow prediction}.
\newblock \bibinfo{journal}{\emph{IEEE Transactions on Knowledge and Data Engineering}} \bibinfo{volume}{35}, \bibinfo{number}{1} (\bibinfo{year}{2021}), \bibinfo{pages}{276--290}.
\newblock


\bibitem[Jiang et~al\mbox{.}(2021b)]%
        {jiang2021dl}
\bibfield{author}{\bibinfo{person}{Renhe Jiang}, \bibinfo{person}{Du Yin}, \bibinfo{person}{Zhaonan Wang}, \bibinfo{person}{Yizhuo Wang}, \bibinfo{person}{Jiewen Deng}, \bibinfo{person}{Hangchen Liu}, \bibinfo{person}{Zekun Cai}, \bibinfo{person}{Jinliang Deng}, \bibinfo{person}{Xuan Song}, {and} \bibinfo{person}{Ryosuke Shibasaki}.} \bibinfo{year}{2021}\natexlab{b}.
\newblock \showarticletitle{Dl-traff: Survey and benchmark of deep learning models for urban traffic prediction}. In \bibinfo{booktitle}{\emph{Proceedings of the ACM International Conference on Information and Knowledge Management}}. \bibinfo{pages}{4515--4525}.
\newblock


\bibitem[Jin et~al\mbox{.}(2024)]%
        {jin2023time}
\bibfield{author}{\bibinfo{person}{Ming Jin}, \bibinfo{person}{Shiyu Wang}, \bibinfo{person}{Lintao Ma}, \bibinfo{person}{Zhixuan Chu}, \bibinfo{person}{James~Y Zhang}, \bibinfo{person}{Xiaoming Shi}, \bibinfo{person}{Pin-Yu Chen}, \bibinfo{person}{Yuxuan Liang}, \bibinfo{person}{Yuan-Fang Li}, \bibinfo{person}{Shirui Pan}, {and} \bibinfo{person}{Qingsong Wen}.} \bibinfo{year}{2024}\natexlab{}.
\newblock \showarticletitle{{Time-LLM}: Time series forecasting by reprogramming large language models}. In \bibinfo{booktitle}{\emph{International Conference on Learning Representations}}.
\newblock


\bibitem[Lee and Ko(2024)]%
        {lee2024testam}
\bibfield{author}{\bibinfo{person}{Hyunwook Lee} {and} \bibinfo{person}{Sungahn Ko}.} \bibinfo{year}{2024}\natexlab{}.
\newblock \showarticletitle{{TESTAM}: A Time-Enhanced Spatio-Temporal Attention Model with Mixture of Experts}. In \bibinfo{booktitle}{\emph{International Conference on Learning Representations}}.
\newblock


\bibitem[Lee et~al\mbox{.}(2019)]%
        {lee2019set}
\bibfield{author}{\bibinfo{person}{Juho Lee}, \bibinfo{person}{Yoonho Lee}, \bibinfo{person}{Jungtaek Kim}, \bibinfo{person}{Adam Kosiorek}, \bibinfo{person}{Seungjin Choi}, {and} \bibinfo{person}{Yee~Whye Teh}.} \bibinfo{year}{2019}\natexlab{}.
\newblock \showarticletitle{Set Transformer: A Framework for Attention-based Permutation-Invariant Neural Networks}. In \bibinfo{booktitle}{\emph{International Conference on Machine Learning}}. \bibinfo{pages}{3744--3753}.
\newblock


\bibitem[Li et~al\mbox{.}(2018)]%
        {li2017diffusion}
\bibfield{author}{\bibinfo{person}{Yaguang Li}, \bibinfo{person}{Rose Yu}, \bibinfo{person}{Cyrus Shahabi}, {and} \bibinfo{person}{Yan Liu}.} \bibinfo{year}{2018}\natexlab{}.
\newblock \showarticletitle{Diffusion Convolutional Recurrent Neural Network: Data-Driven Traffic Forecasting}. In \bibinfo{booktitle}{\emph{International Conference on Learning Representations}}.
\newblock


\bibitem[Lin et~al\mbox{.}(2024)]%
        {lin2024cyclenet}
\bibfield{author}{\bibinfo{person}{Shengsheng Lin}, \bibinfo{person}{Weiwei Lin}, \bibinfo{person}{Xinyi Hu}, \bibinfo{person}{Wentai Wu}, \bibinfo{person}{Ruichao Mo}, {and} \bibinfo{person}{Haocheng Zhong}.} \bibinfo{year}{2024}\natexlab{}.
\newblock \showarticletitle{Cyclenet: Enhancing time series forecasting through modeling periodic patterns}.
\newblock \bibinfo{journal}{\emph{Advances in Neural Information Processing Systems}}  \bibinfo{volume}{37} (\bibinfo{year}{2024}), \bibinfo{pages}{106315--106345}.
\newblock


\bibitem[Liu et~al\mbox{.}(2023)]%
        {liu2023spatio}
\bibfield{author}{\bibinfo{person}{Hangchen Liu}, \bibinfo{person}{Zheng Dong}, \bibinfo{person}{Renhe Jiang}, \bibinfo{person}{Jiewen Deng}, \bibinfo{person}{Jinliang Deng}, \bibinfo{person}{Quanjun Chen}, {and} \bibinfo{person}{Xuan Song}.} \bibinfo{year}{2023}\natexlab{}.
\newblock \showarticletitle{Spatio-temporal adaptive embedding makes vanilla transformer sota for traffic forecasting}. In \bibinfo{booktitle}{\emph{Proceedings of the ACM International Conference on Information and Knowledge management}}. \bibinfo{pages}{4125--4129}.
\newblock


\bibitem[Liu et~al\mbox{.}(2021)]%
        {liu2021community}
\bibfield{author}{\bibinfo{person}{Hao Liu}, \bibinfo{person}{Qiyu Wu}, \bibinfo{person}{Fuzhen Zhuang}, \bibinfo{person}{Xinjiang Lu}, \bibinfo{person}{Dejing Dou}, {and} \bibinfo{person}{Hui Xiong}.} \bibinfo{year}{2021}\natexlab{}.
\newblock \showarticletitle{Community-aware multi-task transportation demand prediction}. In \bibinfo{booktitle}{\emph{Proceedings of the AAAI Conference on Artificial Intelligence}}, Vol.~\bibinfo{volume}{35}. \bibinfo{pages}{320--327}.
\newblock


\bibitem[Liu et~al\mbox{.}(2025)]%
        {liu2025calf}
\bibfield{author}{\bibinfo{person}{Peiyuan Liu}, \bibinfo{person}{Hang Guo}, \bibinfo{person}{Tao Dai}, \bibinfo{person}{Naiqi Li}, \bibinfo{person}{Jigang Bao}, \bibinfo{person}{Xudong Ren}, \bibinfo{person}{Yong Jiang}, {and} \bibinfo{person}{Shu-Tao Xia}.} \bibinfo{year}{2025}\natexlab{}.
\newblock \showarticletitle{Calf: Aligning llms for time series forecasting via cross-modal fine-tuning}. In \bibinfo{booktitle}{\emph{Proceedings of the AAAI Conference on Artificial Intelligence}}, Vol.~\bibinfo{volume}{39}. \bibinfo{pages}{18915--18923}.
\newblock


\bibitem[Liu and Paparrizos(2024)]%
        {liu2024elephant}
\bibfield{author}{\bibinfo{person}{Qinghua Liu} {and} \bibinfo{person}{John Paparrizos}.} \bibinfo{year}{2024}\natexlab{}.
\newblock \showarticletitle{The elephant in the room: Towards a reliable time-series anomaly detection benchmark}.
\newblock \bibinfo{journal}{\emph{Advances in Neural Information Processing Systems}}  \bibinfo{volume}{37} (\bibinfo{year}{2024}), \bibinfo{pages}{108231--108261}.
\newblock


\bibitem[Liu et~al\mbox{.}(2024)]%
        {liu2024itransformer}
\bibfield{author}{\bibinfo{person}{Yong Liu}, \bibinfo{person}{Tengge Hu}, \bibinfo{person}{Haoran Zhang}, \bibinfo{person}{Haixu Wu}, \bibinfo{person}{Shiyu Wang}, \bibinfo{person}{Lintao Ma}, {and} \bibinfo{person}{Mingsheng Long}.} \bibinfo{year}{2024}\natexlab{}.
\newblock \showarticletitle{iTransformer: Inverted Transformers Are Effective for Time Series Forecasting}. In \bibinfo{booktitle}{\emph{International Conference on Learning Representations}}.
\newblock


\bibitem[Lu et~al\mbox{.}(2023)]%
        {lu2023outofdistribution}
\bibfield{author}{\bibinfo{person}{Wang Lu}, \bibinfo{person}{Jindong Wang}, \bibinfo{person}{Xinwei Sun}, \bibinfo{person}{Yiqiang Chen}, {and} \bibinfo{person}{Xing Xie}.} \bibinfo{year}{2023}\natexlab{}.
\newblock \showarticletitle{Out-of-distribution Representation Learning for Time Series Classification}. In \bibinfo{booktitle}{\emph{International Conference on Learning Representations}}.
\newblock


\bibitem[Nie et~al\mbox{.}(2023)]%
        {Yuqietal2023PatchTST}
\bibfield{author}{\bibinfo{person}{Yuqi Nie}, \bibinfo{person}{Nam H.~Nguyen}, \bibinfo{person}{Phanwadee Sinthong}, {and} \bibinfo{person}{Jayant Kalagnanam}.} \bibinfo{year}{2023}\natexlab{}.
\newblock \showarticletitle{A Time Series is Worth 64 Words: Long-term Forecasting with Transformers}. In \bibinfo{booktitle}{\emph{International Conference on Learning Representations}}.
\newblock


\bibitem[Oreshkin et~al\mbox{.}(2021)]%
        {oreshkin2021fc}
\bibfield{author}{\bibinfo{person}{Boris~N Oreshkin}, \bibinfo{person}{Arezou Amini}, \bibinfo{person}{Lucy Coyle}, {and} \bibinfo{person}{Mark Coates}.} \bibinfo{year}{2021}\natexlab{}.
\newblock \showarticletitle{FC-GAGA: Fully connected gated graph architecture for spatio-temporal traffic forecasting}. In \bibinfo{booktitle}{\emph{Proceedings of the AAAI Conference on Artificial Intelligence}}, Vol.~\bibinfo{volume}{35}. \bibinfo{pages}{9233--9241}.
\newblock


\bibitem[Pan et~al\mbox{.}(2019)]%
        {pan2019urban}
\bibfield{author}{\bibinfo{person}{Zheyi Pan}, \bibinfo{person}{Yuxuan Liang}, \bibinfo{person}{Weifeng Wang}, \bibinfo{person}{Yong Yu}, \bibinfo{person}{Yu Zheng}, {and} \bibinfo{person}{Junbo Zhang}.} \bibinfo{year}{2019}\natexlab{}.
\newblock \showarticletitle{Urban traffic prediction from spatio-temporal data using deep meta learning}. In \bibinfo{booktitle}{\emph{Proceedings of the ACM SIGKDD Conference on Knowledge Discovery and Data Mining}}. \bibinfo{pages}{1720--1730}.
\newblock


\bibitem[Piao et~al\mbox{.}(2024)]%
        {piao2024fredformer}
\bibfield{author}{\bibinfo{person}{Xihao Piao}, \bibinfo{person}{Zheng Chen}, \bibinfo{person}{Taichi Murayama}, \bibinfo{person}{Yasuko Matsubara}, {and} \bibinfo{person}{Yasushi Sakurai}.} \bibinfo{year}{2024}\natexlab{}.
\newblock \showarticletitle{Fredformer: Frequency debiased transformer for time series forecasting}. In \bibinfo{booktitle}{\emph{Proceedings of the ACM SIGKDD Conference on Knowledge Discovery and Data Mining}}. \bibinfo{pages}{2400--2410}.
\newblock


\bibitem[Qiu et~al\mbox{.}(2025)]%
        {qiu2025duet}
\bibfield{author}{\bibinfo{person}{Xiangfei Qiu}, \bibinfo{person}{Xingjian Wu}, \bibinfo{person}{Yan Lin}, \bibinfo{person}{Chenjuan Guo}, \bibinfo{person}{Jilin Hu}, {and} \bibinfo{person}{Bin Yang}.} \bibinfo{year}{2025}\natexlab{}.
\newblock \showarticletitle{Duet: Dual clustering enhanced multivariate time series forecasting}. In \bibinfo{booktitle}{\emph{Proceedings of the ACM SIGKDD Conference on Knowledge Discovery and Data Mining}}. \bibinfo{pages}{1185--1196}.
\newblock


\bibitem[Shao et~al\mbox{.}(2022)]%
        {shao2022pre}
\bibfield{author}{\bibinfo{person}{Zezhi Shao}, \bibinfo{person}{Zhao Zhang}, \bibinfo{person}{Fei Wang}, {and} \bibinfo{person}{Yongjun Xu}.} \bibinfo{year}{2022}\natexlab{}.
\newblock \showarticletitle{Pre-training enhanced spatial-temporal graph neural network for multivariate time series forecasting}. In \bibinfo{booktitle}{\emph{Proceedings of the ACM SIGKDD Conference on Knowledge Discovery and Data Mining}}. \bibinfo{pages}{1567--1577}.
\newblock


\bibitem[van~den Oord et~al\mbox{.}(2016)]%
        {Oord2016WaveNetAG}
\bibfield{author}{\bibinfo{person}{A{\"a}ron van~den Oord}, \bibinfo{person}{Sander Dieleman}, \bibinfo{person}{Heiga Zen}, \bibinfo{person}{Karen Simonyan}, \bibinfo{person}{Oriol Vinyals}, \bibinfo{person}{Alex Graves}, \bibinfo{person}{Nal Kalchbrenner}, \bibinfo{person}{Andrew~W. Senior}, {and} \bibinfo{person}{Koray Kavukcuoglu}.} \bibinfo{year}{2016}\natexlab{}.
\newblock \showarticletitle{WaveNet: A Generative Model for Raw Audio}. In \bibinfo{booktitle}{\emph{Proceedings of the ISCA Workshop on Speech Synthesis Workshop}}.
\newblock


\bibitem[Vaswani et~al\mbox{.}(2017)]%
        {vaswani2017attention}
\bibfield{author}{\bibinfo{person}{Ashish Vaswani}, \bibinfo{person}{Noam Shazeer}, \bibinfo{person}{Niki Parmar}, \bibinfo{person}{Jakob Uszkoreit}, \bibinfo{person}{Llion Jones}, \bibinfo{person}{Aidan~N Gomez}, \bibinfo{person}{{\L}ukasz Kaiser}, {and} \bibinfo{person}{Illia Polosukhin}.} \bibinfo{year}{2017}\natexlab{}.
\newblock \showarticletitle{Attention is all you need}.
\newblock \bibinfo{journal}{\emph{Advances in Neural Information Processing Systems}}  \bibinfo{volume}{30} (\bibinfo{year}{2017}).
\newblock


\bibitem[Wang et~al\mbox{.}(2023)]%
        {wang2023drift}
\bibfield{author}{\bibinfo{person}{Chengsen Wang}, \bibinfo{person}{Zirui Zhuang}, \bibinfo{person}{Qi Qi}, \bibinfo{person}{Jingyu Wang}, \bibinfo{person}{Xingyu Wang}, \bibinfo{person}{Haifeng Sun}, {and} \bibinfo{person}{Jianxin Liao}.} \bibinfo{year}{2023}\natexlab{}.
\newblock \showarticletitle{Drift doesn't matter: Dynamic decomposition with diffusion reconstruction for unstable multivariate time series anomaly detection}.
\newblock \bibinfo{journal}{\emph{Advances in Neural Information Processing Systems}}  \bibinfo{volume}{36} (\bibinfo{year}{2023}), \bibinfo{pages}{10758--10774}.
\newblock


\bibitem[Wang et~al\mbox{.}(2025)]%
        {wang2025timemixer++}
\bibfield{author}{\bibinfo{person}{Shiyu Wang}, \bibinfo{person}{Jiawei Li}, \bibinfo{person}{Xiaoming Shi}, \bibinfo{person}{Zhou Ye}, \bibinfo{person}{Baichuan Mo}, \bibinfo{person}{Wenze Lin}, \bibinfo{person}{Shengtong Ju}, \bibinfo{person}{Zhixuan Chu}, {and} \bibinfo{person}{Ming Jin}.} \bibinfo{year}{2025}\natexlab{}.
\newblock \showarticletitle{TimeMixer++: A General Time Series Pattern Machine for Universal Predictive Analysis}. In \bibinfo{booktitle}{\emph{International Conference on Learning Representations}}.
\newblock


\bibitem[Wang et~al\mbox{.}(2024)]%
        {wang2023timemixer}
\bibfield{author}{\bibinfo{person}{Shiyu Wang}, \bibinfo{person}{Haixu Wu}, \bibinfo{person}{Xiaoming Shi}, \bibinfo{person}{Tengge Hu}, \bibinfo{person}{Huakun Luo}, \bibinfo{person}{Lintao Ma}, \bibinfo{person}{James~Y Zhang}, {and} \bibinfo{person}{JUN ZHOU}.} \bibinfo{year}{2024}\natexlab{}.
\newblock \showarticletitle{TimeMixer: Decomposable Multiscale Mixing for Time Series Forecasting}. In \bibinfo{booktitle}{\emph{International Conference on Learning Representations}}.
\newblock


\bibitem[Wang et~al\mbox{.}(2022)]%
        {wang2022event}
\bibfield{author}{\bibinfo{person}{Zhaonan Wang}, \bibinfo{person}{Renhe Jiang}, \bibinfo{person}{Hao Xue}, \bibinfo{person}{Flora~D Salim}, \bibinfo{person}{Xuan Song}, {and} \bibinfo{person}{Ryosuke Shibasaki}.} \bibinfo{year}{2022}\natexlab{}.
\newblock \showarticletitle{Event-aware multimodal mobility nowcasting}. In \bibinfo{booktitle}{\emph{Proceedings of the AAAI Conference on Artificial Intelligence}}, Vol.~\bibinfo{volume}{36}. \bibinfo{pages}{4228--4236}.
\newblock


\bibitem[Wu et~al\mbox{.}(2023)]%
        {wu2023timesnet}
\bibfield{author}{\bibinfo{person}{Haixu Wu}, \bibinfo{person}{Tengge Hu}, \bibinfo{person}{Yong Liu}, \bibinfo{person}{Hang Zhou}, \bibinfo{person}{Jianmin Wang}, {and} \bibinfo{person}{Mingsheng Long}.} \bibinfo{year}{2023}\natexlab{}.
\newblock \showarticletitle{TimesNet: Temporal 2D-Variation Modeling for General Time Series Analysis}. In \bibinfo{booktitle}{\emph{International Conference on Learning Representations}}.
\newblock


\bibitem[Wu et~al\mbox{.}(2021)]%
        {wu2021autoformer}
\bibfield{author}{\bibinfo{person}{Haixu Wu}, \bibinfo{person}{Jiehui Xu}, \bibinfo{person}{Jianmin Wang}, {and} \bibinfo{person}{Mingsheng Long}.} \bibinfo{year}{2021}\natexlab{}.
\newblock \showarticletitle{Autoformer: Decomposition transformers with auto-correlation for long-term series forecasting}.
\newblock \bibinfo{journal}{\emph{Advances in Neural Information Processing Systems}}  \bibinfo{volume}{34} (\bibinfo{year}{2021}), \bibinfo{pages}{22419--22430}.
\newblock


\bibitem[Wu et~al\mbox{.}(2020a)]%
        {wu2020hierarchically}
\bibfield{author}{\bibinfo{person}{Xian Wu}, \bibinfo{person}{Chao Huang}, \bibinfo{person}{Chuxu Zhang}, {and} \bibinfo{person}{Nitesh~V Chawla}.} \bibinfo{year}{2020}\natexlab{a}.
\newblock \showarticletitle{Hierarchically structured transformer networks for fine-grained spatial event forecasting}. In \bibinfo{booktitle}{\emph{Proceedings of the Web Conference}}. \bibinfo{pages}{2320--2330}.
\newblock


\bibitem[Wu et~al\mbox{.}(2020b)]%
        {wu2020connecting}
\bibfield{author}{\bibinfo{person}{Zonghan Wu}, \bibinfo{person}{Shirui Pan}, \bibinfo{person}{Guodong Long}, \bibinfo{person}{Jing Jiang}, \bibinfo{person}{Xiaojun Chang}, {and} \bibinfo{person}{Chengqi Zhang}.} \bibinfo{year}{2020}\natexlab{b}.
\newblock \showarticletitle{Connecting the dots: Multivariate time series forecasting with graph neural networks}. In \bibinfo{booktitle}{\emph{Proceedings of the ACM SIGKDD Conference on Knowledge Discovery and Data Mining}}. \bibinfo{pages}{753--763}.
\newblock


\bibitem[Wu et~al\mbox{.}(2019)]%
        {wu2019graph}
\bibfield{author}{\bibinfo{person}{Zonghan Wu}, \bibinfo{person}{Shirui Pan}, \bibinfo{person}{Guodong Long}, \bibinfo{person}{Jing Jiang}, {and} \bibinfo{person}{Chengqi Zhang}.} \bibinfo{year}{2019}\natexlab{}.
\newblock \showarticletitle{Graph wavenet for deep spatial-temporal graph modeling}. In \bibinfo{booktitle}{\emph{Proceedings of the International Joint Conference on Artificial Intelligence}}. \bibinfo{pages}{1907--1913}.
\newblock


\bibitem[Xia et~al\mbox{.}(2021)]%
        {xia2021stshn}
\bibfield{author}{\bibinfo{person}{Lianghao Xia}, \bibinfo{person}{Chao Huang}, \bibinfo{person}{Yong Xu}, \bibinfo{person}{Peng Dai}, \bibinfo{person}{Liefeng Bo}, \bibinfo{person}{Xiyue Zhang}, {and} \bibinfo{person}{Tianyi Chen}.} \bibinfo{year}{2021}\natexlab{}.
\newblock \showarticletitle{Spatial-Temporal Sequential Hypergraph Network for Crime Prediction with Dynamic Multiplex Relation Learning}. In \bibinfo{booktitle}{\emph{Proceedings of the International Joint Conference on Artificial Intelligence}}. \bibinfo{pages}{1631--1637}.
\newblock


\bibitem[Yang et~al\mbox{.}(2021)]%
        {yang2021space}
\bibfield{author}{\bibinfo{person}{Song Yang}, \bibinfo{person}{Jiamou Liu}, {and} \bibinfo{person}{Kaiqi Zhao}.} \bibinfo{year}{2021}\natexlab{}.
\newblock \showarticletitle{Space Meets Time: Local Spacetime Neural Network For Traffic Flow Forecasting}. In \bibinfo{booktitle}{\emph{IEEE International Conference on Data Mining}}. \bibinfo{pages}{817--826}.
\newblock


\bibitem[Yao et~al\mbox{.}(2019)]%
        {yao2019learning}
\bibfield{author}{\bibinfo{person}{Huaxiu Yao}, \bibinfo{person}{Yiding Liu}, \bibinfo{person}{Ying Wei}, \bibinfo{person}{Xianfeng Tang}, {and} \bibinfo{person}{Zhenhui Li}.} \bibinfo{year}{2019}\natexlab{}.
\newblock \showarticletitle{Learning from multiple cities: A meta-learning approach for spatial-temporal prediction}. In \bibinfo{booktitle}{\emph{The World Wide Web conference}}. \bibinfo{pages}{2181--2191}.
\newblock


\bibitem[Ye et~al\mbox{.}(2019)]%
        {ye2019co}
\bibfield{author}{\bibinfo{person}{Junchen Ye}, \bibinfo{person}{Leilei Sun}, \bibinfo{person}{Bowen Du}, \bibinfo{person}{Yanjie Fu}, \bibinfo{person}{Xinran Tong}, {and} \bibinfo{person}{Hui Xiong}.} \bibinfo{year}{2019}\natexlab{}.
\newblock \showarticletitle{Co-prediction of multiple transportation demands based on deep spatio-temporal neural network}. In \bibinfo{booktitle}{\emph{Proceedings of the ACM SIGKDD Conference on Knowledge Discovery and Data Mining}}. \bibinfo{pages}{305--313}.
\newblock


\bibitem[Ye et~al\mbox{.}(2024)]%
        {ye2024frequency}
\bibfield{author}{\bibinfo{person}{Weiwei Ye}, \bibinfo{person}{Songgaojun Deng}, \bibinfo{person}{Qiaosha Zou}, {and} \bibinfo{person}{Ning Gui}.} \bibinfo{year}{2024}\natexlab{}.
\newblock \showarticletitle{Frequency adaptive normalization for non-stationary time series forecasting}.
\newblock \bibinfo{journal}{\emph{Advances in Neural Information Processing Systems}}  \bibinfo{volume}{37} (\bibinfo{year}{2024}), \bibinfo{pages}{31350--31379}.
\newblock


\bibitem[Yi et~al\mbox{.}(2024)]%
        {yi2024filternet}
\bibfield{author}{\bibinfo{person}{Kun Yi}, \bibinfo{person}{Jingru Fei}, \bibinfo{person}{Qi Zhang}, \bibinfo{person}{Hui He}, \bibinfo{person}{Shufeng Hao}, \bibinfo{person}{Defu Lian}, {and} \bibinfo{person}{Wei Fan}.} \bibinfo{year}{2024}\natexlab{}.
\newblock \showarticletitle{Filternet: Harnessing frequency filters for time series forecasting}.
\newblock \bibinfo{journal}{\emph{Advances in Neural Information Processing Systems}}  \bibinfo{volume}{37} (\bibinfo{year}{2024}), \bibinfo{pages}{55115--55140}.
\newblock


\bibitem[Yi et~al\mbox{.}(2023a)]%
        {yi2023fouriergnn}
\bibfield{author}{\bibinfo{person}{Kun Yi}, \bibinfo{person}{Qi Zhang}, \bibinfo{person}{Wei Fan}, \bibinfo{person}{Hui He}, \bibinfo{person}{Liang Hu}, \bibinfo{person}{Pengyang Wang}, \bibinfo{person}{Ning An}, \bibinfo{person}{Longbing Cao}, {and} \bibinfo{person}{Zhendong Niu}.} \bibinfo{year}{2023}\natexlab{a}.
\newblock \showarticletitle{FourierGNN: Rethinking multivariate time series forecasting from a pure graph perspective}.
\newblock \bibinfo{journal}{\emph{Advances in Neural Information Processing Systems}}  \bibinfo{volume}{36} (\bibinfo{year}{2023}), \bibinfo{pages}{69638--69660}.
\newblock


\bibitem[Yi et~al\mbox{.}(2023b)]%
        {yi2023frequency}
\bibfield{author}{\bibinfo{person}{Kun Yi}, \bibinfo{person}{Qi Zhang}, \bibinfo{person}{Wei Fan}, \bibinfo{person}{Shoujin Wang}, \bibinfo{person}{Pengyang Wang}, \bibinfo{person}{Hui He}, \bibinfo{person}{Ning An}, \bibinfo{person}{Defu Lian}, \bibinfo{person}{Longbing Cao}, {and} \bibinfo{person}{Zhendong Niu}.} \bibinfo{year}{2023}\natexlab{b}.
\newblock \showarticletitle{Frequency-domain MLPs are more effective learners in time series forecasting}.
\newblock \bibinfo{journal}{\emph{Advances in Neural Information Processing Systems}}  \bibinfo{volume}{36} (\bibinfo{year}{2023}), \bibinfo{pages}{76656--76679}.
\newblock


\bibitem[Yu et~al\mbox{.}(2024)]%
        {yu2024ginar}
\bibfield{author}{\bibinfo{person}{Chengqing Yu}, \bibinfo{person}{Fei Wang}, \bibinfo{person}{Zezhi Shao}, \bibinfo{person}{Tangwen Qian}, \bibinfo{person}{Zhao Zhang}, \bibinfo{person}{Wei Wei}, {and} \bibinfo{person}{Yongjun Xu}.} \bibinfo{year}{2024}\natexlab{}.
\newblock \showarticletitle{Ginar: An end-to-end multivariate time series forecasting model suitable for variable missing}. In \bibinfo{booktitle}{\emph{Proceedings of the ACM SIGKDD Conference on Knowledge Discovery and Data Mining}}. \bibinfo{pages}{3989--4000}.
\newblock


\bibitem[Yuan et~al\mbox{.}(2021)]%
        {yuan2021effective}
\bibfield{author}{\bibinfo{person}{Haitao Yuan}, \bibinfo{person}{Guoliang Li}, \bibinfo{person}{Zhifeng Bao}, {and} \bibinfo{person}{Ling Feng}.} \bibinfo{year}{2021}\natexlab{}.
\newblock \showarticletitle{An effective joint prediction model for travel demands and traffic flows}. In \bibinfo{booktitle}{\emph{IEEE International Conference on Data Engineering}}. \bibinfo{pages}{348--359}.
\newblock


\bibitem[Yue et~al\mbox{.}(2025)]%
        {yue2025freeformer}
\bibfield{author}{\bibinfo{person}{Wenzhen Yue}, \bibinfo{person}{Yong Liu}, \bibinfo{person}{Xianghua Ying}, \bibinfo{person}{Bowei Xing}, \bibinfo{person}{Ruohao Guo}, {and} \bibinfo{person}{Ji Shi}.} \bibinfo{year}{2025}\natexlab{}.
\newblock \showarticletitle{FreEformer: frequency enhanced transformer for multivariate time series forecasting}. In \bibinfo{booktitle}{\emph{Proceedings of the International Joint Conference on Artificial Intelligence}}. \bibinfo{pages}{3606--3614}.
\newblock


\bibitem[Zhang et~al\mbox{.}(2024)]%
        {zhang2024multimodal}
\bibfield{author}{\bibinfo{person}{Dongran Zhang}, \bibinfo{person}{Jiangnan Yan}, \bibinfo{person}{Kemal Polat}, \bibinfo{person}{Adi Alhudhaif}, {and} \bibinfo{person}{Jun Li}.} \bibinfo{year}{2024}\natexlab{}.
\newblock \showarticletitle{Multimodal joint prediction of traffic spatial-temporal data with graph sparse attention mechanism and bidirectional temporal convolutional network}.
\newblock \bibinfo{journal}{\emph{Advanced Engineering Informatics}}  \bibinfo{volume}{62} (\bibinfo{year}{2024}), \bibinfo{pages}{102533}.
\newblock


\bibitem[Zhang et~al\mbox{.}(2017)]%
        {zhang2017stresnet}
\bibfield{author}{\bibinfo{person}{Junbo Zhang}, \bibinfo{person}{Yu Zheng}, {and} \bibinfo{person}{Dekang Qi}.} \bibinfo{year}{2017}\natexlab{}.
\newblock \showarticletitle{Deep spatio-temporal residual networks for citywide crowd flows prediction}. In \bibinfo{booktitle}{\emph{Proceedings of the AAAI Conference on Artificial Intelligence}}, Vol.~\bibinfo{volume}{31}. \bibinfo{pages}{1655–1661}.
\newblock


\bibitem[Zhang et~al\mbox{.}(2023)]%
        {zhang2023autost}
\bibfield{author}{\bibinfo{person}{Qianru Zhang}, \bibinfo{person}{Chao Huang}, \bibinfo{person}{Lianghao Xia}, \bibinfo{person}{Zheng Wang}, \bibinfo{person}{Zhonghang Li}, {and} \bibinfo{person}{Siuming Yiu}.} \bibinfo{year}{2023}\natexlab{}.
\newblock \showarticletitle{Automated Spatio-Temporal Graph Contrastive Learning}. In \bibinfo{booktitle}{\emph{Proceedings of the ACM Web Conference}}. \bibinfo{pages}{295--305}.
\newblock


\bibitem[Zhang et~al\mbox{.}(2022)]%
        {zhang2022self}
\bibfield{author}{\bibinfo{person}{Xiang Zhang}, \bibinfo{person}{Ziyuan Zhao}, \bibinfo{person}{Theodoros Tsiligkaridis}, {and} \bibinfo{person}{Marinka Zitnik}.} \bibinfo{year}{2022}\natexlab{}.
\newblock \showarticletitle{Self-supervised contrastive pre-training for time series via time-frequency consistency}.
\newblock \bibinfo{journal}{\emph{Advances in Neural Information Processing Systems}}  \bibinfo{volume}{35} (\bibinfo{year}{2022}), \bibinfo{pages}{3988--4003}.
\newblock


\bibitem[Zhang and Yan(2023)]%
        {zhang2023crossformer}
\bibfield{author}{\bibinfo{person}{Yunhao Zhang} {and} \bibinfo{person}{Junchi Yan}.} \bibinfo{year}{2023}\natexlab{}.
\newblock \showarticletitle{Crossformer: Transformer Utilizing Cross-Dimension Dependency for Multivariate Time Series Forecasting}. In \bibinfo{booktitle}{\emph{International Conference on Learning Representations}}.
\newblock


\bibitem[Zhao and Shen(2024)]%
        {zhao2024lift}
\bibfield{author}{\bibinfo{person}{Lifan Zhao} {and} \bibinfo{person}{Yanyan Shen}.} \bibinfo{year}{2024}\natexlab{}.
\newblock \showarticletitle{Rethinking Channel Dependence for Multivariate Time Series Forecasting: Learning from Leading Indicators}. In \bibinfo{booktitle}{\emph{International Conference on Learning Representations}}.
\newblock


\bibitem[Zheng et~al\mbox{.}(2020)]%
        {zheng2020gman}
\bibfield{author}{\bibinfo{person}{Chuanpan Zheng}, \bibinfo{person}{Xiaoliang Fan}, \bibinfo{person}{Cheng Wang}, {and} \bibinfo{person}{Jianzhong Qi}.} \bibinfo{year}{2020}\natexlab{}.
\newblock \showarticletitle{Gman: A graph multi-attention network for traffic prediction}. In \bibinfo{booktitle}{\emph{Proceedings of the AAAI conference on artificial intelligence}}, Vol.~\bibinfo{volume}{34}. \bibinfo{pages}{1234--1241}.
\newblock


\bibitem[Zhou et~al\mbox{.}(2021)]%
        {zhou2021informer}
\bibfield{author}{\bibinfo{person}{Haoyi Zhou}, \bibinfo{person}{Shanghang Zhang}, \bibinfo{person}{Jieqi Peng}, \bibinfo{person}{Shuai Zhang}, \bibinfo{person}{Jianxin Li}, \bibinfo{person}{Hui Xiong}, {and} \bibinfo{person}{Wancai Zhang}.} \bibinfo{year}{2021}\natexlab{}.
\newblock \showarticletitle{Informer: Beyond efficient transformer for long sequence time-series forecasting}. In \bibinfo{booktitle}{\emph{Proceedings of the AAAI Conference on Artificial Intelligence}}, Vol.~\bibinfo{volume}{35}. \bibinfo{pages}{11106--11115}.
\newblock


\bibitem[Zhou et~al\mbox{.}(2022)]%
        {zhou2022fedformer}
\bibfield{author}{\bibinfo{person}{Tian Zhou}, \bibinfo{person}{Ziqing Ma}, \bibinfo{person}{Qingsong Wen}, \bibinfo{person}{Xue Wang}, \bibinfo{person}{Liang Sun}, {and} \bibinfo{person}{Rong Jin}.} \bibinfo{year}{2022}\natexlab{}.
\newblock \showarticletitle{Fedformer: Frequency enhanced decomposed transformer for long-term series forecasting}. In \bibinfo{booktitle}{\emph{International Conference on Machine Learning}}. \bibinfo{pages}{27268--27286}.
\newblock


\bibitem[Zhou et~al\mbox{.}(2023)]%
        {zhou2023one}
\bibfield{author}{\bibinfo{person}{Tian Zhou}, \bibinfo{person}{Peisong Niu}, \bibinfo{person}{Liang Sun}, \bibinfo{person}{Rong Jin}, {et~al\mbox{.}}} \bibinfo{year}{2023}\natexlab{}.
\newblock \showarticletitle{One fits all: Power general time series analysis by pretrained lm}.
\newblock \bibinfo{journal}{\emph{Advances in neural information processing systems}}  \bibinfo{volume}{36} (\bibinfo{year}{2023}), \bibinfo{pages}{43322--43355}.
\newblock


\bibitem[Zhou et~al\mbox{.}(2018)]%
        {zhou2018attconvlstm}
\bibfield{author}{\bibinfo{person}{Xian Zhou}, \bibinfo{person}{Yanyan Shen}, \bibinfo{person}{Yanmin Zhu}, {and} \bibinfo{person}{Linpeng Huang}.} \bibinfo{year}{2018}\natexlab{}.
\newblock \showarticletitle{Predicting Multi-step Citywide Passenger Demands Using Attention-based Neural Networks}. In \bibinfo{booktitle}{\emph{Proceedings of the ACM International Conference on Web Search and Data Mining}}. \bibinfo{pages}{736--744}.
\newblock


\bibitem[Zhu et~al\mbox{.}(2021)]%
        {zhu2021mixseq}
\bibfield{author}{\bibinfo{person}{Zhibo Zhu}, \bibinfo{person}{Ziqi Liu}, \bibinfo{person}{Ge Jin}, \bibinfo{person}{Zhiqiang Zhang}, \bibinfo{person}{Lei Chen}, \bibinfo{person}{Jun Zhou}, {and} \bibinfo{person}{Jianyong Zhou}.} \bibinfo{year}{2021}\natexlab{}.
\newblock \showarticletitle{MixSeq: Connecting Macroscopic Time Series Forecasting with Microscopic Time Series Data}.
\newblock \bibinfo{journal}{\emph{Advances in Neural Information Processing Systems}}  \bibinfo{volume}{34} (\bibinfo{year}{2021}), \bibinfo{pages}{12904--12916}.
\newblock


\end{thebibliography}

\appendix
\balance

\section{Complexity and Efficiency Analysis}
\noindent \textbf{Theoretical Complexity.}
FreMo consists of three main costs. (1) rFFT and irFFT along the temporal dimension cost $\mathcal{O}(MNd \cdot T \log T)$, which is the only term super-linear in $T$. (2) MFF performs amplitude computation, channel pooling, frequency-wise gating, and gate generation, all linear in $F$, with cost $\mathcal{O}(MNd \cdot F)$. (3) FSI performs frequency-wise scoring, modality Softmax, consensus aggregation, and residual feedback, with cost $\mathcal{O}(MNd \cdot F + MNF)$, where $\mathcal{O}(MNF)$ comes from the Softmax. Thus, the total time complexity is $\mathcal{O}(MNd \cdot T \log T + MNd \cdot F)$, which asymptotically reduces to $\mathcal{O}(MNd \cdot T \log T)$.

\noindent \textbf{Practical Efficiency.}
FreMo processes modalities and nodes in parallel, with costs scaling linearly with $M$ and $N$. Its additional memory footprint is $\mathcal{O}(MNFd)$ due to intermediate frequency-domain tensors.  As shown in Table~\ref{tab:overhead}, FreMo adds only $+0.028$M parameters on NYC/DC and $+0.054$M on Chicago, with moderate training-time overhead ranging from $+1.2$s to $+13.9$s per epoch. This confirms that FreMo is lightweight, highly parallelizable, and can be efficiently integrated into diverse forecasting backbones for large-scale multi-modality transportation forecasting.

\begin{table}[H]
\centering
\caption{Overhead of Params / Training Time.}
\setlength{\tabcolsep}{1.5pt}
\resizebox{\linewidth}{!}{%
    \begin{tabular}{c|c|c|c}
    \toprule    
    
    \textbf{Model} & \textbf{NYC} & \textbf{DC} & \textbf{Chicago} \\
    \midrule
    
    AGCRN~\cite{bai2020adaptive} &
    0.597M / 48.9s & 0.597M / 27.3s & 0.600M / 90.1s \\
    \textbf{+FreMo} &
    +0.028M / +2.4s & +0.028M / +10.6s & +0.054M / +13.9s \\
    \midrule

    TimesNet~\cite{wu2023timesnet}&
    4.718M / 28.3s & 4.720M / 15.8s & 4.824M / 29.8s \\
    \textbf{+FreMo} &
    +0.028M / +3.1s & +0.028M / +7.9s & +0.054M / +12.8s \\
    \midrule
    
    iTransformer~\cite{liu2024itransformer}&
    0.052M / 7.6s & 0.052M / 4.5s  & 0.052M / 21.8s \\
    \textbf{+FreMo} &
    +0.028M / +4.0s & +0.028M / +3.4s & +0.054M / +12.3s \\
    \midrule

    STAEformer~\cite{liu2023spatio}& 
    1.187M / 16.2s & 1.199M / 23.3s & 1.714M / 101.2s \\
    \textbf{+FreMo} &
    +0.028M / ++1.2s & +0.028M / +5.8s & +0.054M / +12.6s \\
    
    \bottomrule
    \end{tabular}%
}
\label{tab:overhead}
\end{table}

\begin{table*}[t]
\centering
\scriptsize
\caption{Full results of applying FreMo to representative time series models.}
\renewcommand{\arraystretch}{1}
\setlength{\tabcolsep}{2.5pt}
\resizebox{\textwidth}{!}{%
    \begin{tabular}{c|cc|cc|cc|cc}
    \toprule
    \toprule
    \multirow{2}{*}{\textbf{NYC}} & 
    \multicolumn{2}{c|}{Bike Inflow} & 
    \multicolumn{2}{c|}{Bike Outflow} & 
    \multicolumn{2}{c|}{Taxi Inflow} & 
    \multicolumn{2}{c}{Taxi Outflow} \\
    \cmidrule(lr){2-3} \cmidrule(lr){4-5} \cmidrule(lr){6-7} \cmidrule(lr){8-9}
    
     & MAE & RMSE & MAE & RMSE & MAE & RMSE & MAE & RMSE \\
    \midrule
    
    AGCRN~\cite{bai2020adaptive} &
    2.40 / 2.70 / 3.02 & 6.96 / 7.69 / 8.41 & 2.60 / 2.97 / 3.31 & 7.39 / 8.32 / 9.15 & 7.08 / 8.10 / 9.16 & 13.42 / 15.71 / 18.16 & 7.81 / 9.03 / 10.19 & 15.03 / 17.41 / 19.29 \\
    \textbf{+ FreMo} &
    \textbf{2.37 / 2.63 / 2.89} & \textbf{4.91 / 5.31 / 5.78} & \textbf{2.54 / 2.87 / 3.27} & \textbf{5.21 / 5.75 / 6.18} & \textbf{6.86 / 7.92 / 8.84} & \textbf{13.13 / 14.72 / 16.69} & \textbf{7.81 / 8.75 / 9.75} & \textbf{14.63 / 16.34 / 18.07} \\
    \midrule
    
    TimesNet~\cite{wu2023timesnet} &
    2.65 / 2.88 / 3.21 & 4.61 / 5.11 / 5.88  & 2.81 / 3.10 / 3.45 & 5.03 / 5.65 / 6.34 & 7.54 / 8.44 / 9.25 & 13.61 / 15.98 / 18.04 & 8.45 / 9.52 / 10.57 & 15.94 / 18.15 / 20.37 \\
    \textbf{+ FreMo} &
    \textbf{2.65 / 2.86 / 3.15} & \textbf{4.55 / 4.98 / 5.78} & \textbf{2.79 / 3.07 / 3.41} & \textbf{4.94 / 5.54 / 6.26} & \textbf{7.50 / 8.40 / 9.22} & \textbf{13.57 / 15.65 / 17.95} & \textbf{8.42 / 9.43 / 10.35} & \textbf{15.93 / 18.06 / 20.02} \\
    \midrule
    
    iTransformer~\cite{liu2024itransformer} &
    3.05 / 3.75 / 4.52 & 5.61 / 7.22 / 8.85 & 3.31 / 4.10 / 4.80 & 6.20 / 7.99 / 9.44 & 8.11 / 10.32 / 12.54 & 14.40 / 19.16 / 23.98 & 8.69 / 10.85 / 12.80 & 16.15 / 20.60 / 24.37 \\
    \textbf{+ FreMo} &
    \textbf{2.93 / 3.36 / 3.77} & \textbf{5.20 / 6.06 / 6.88} & \textbf{3.09 / 3.54 / 3.94} & \textbf{5.79 / 6.73 / 7.58} & \textbf{7.71 / 9.21 / 10.66} & \textbf{13.47 / 16.73 / 19.81} & \textbf{8.44 / 9.91 / 11.38} & \textbf{15.46 / 18.33 / 20.75} \\
    \midrule

    STAEformer~\cite{liu2023spatio} & 3.57 / 3.67 / 3.81 & 7.13 / 7.42 / 7.78 & 3.44 / 3.57 / 3.73 & 7.10 / 7.40 / 7.74 & 8.76 / 10.33 / 11.89 & 15.40 / 19.51 / 24.06 & 9.68 / 10.96 / 12.14 & 16.74 / 19.34 / 21.75 \\
    
    \textbf{+ FreMo} &
    \textbf{2.62} / \textbf{2.83} / \textbf{3.10} & \textbf{5.01} / \textbf{5.58} / \textbf{6.32} & \textbf{2.72} / \textbf{2.99} / \textbf{3.24} & \textbf{5.48} / 6.29 / \textbf{7.07} & \textbf{7.24} / \textbf{8.19} / \textbf{9.26} & \textbf{13.92} / \textbf{16.07} / \textbf{19.02} & \textbf{7.69} / \textbf{8.60} / \textbf{9.46} & \textbf{14.77} / \textbf{16.76} / \textbf{18.48} \\
    \midrule
    \midrule
    \multirow{2}{*}{\textbf{DC}} & 
    \multicolumn{2}{c}{Bike Inflow} & 
    \multicolumn{2}{c}{Bike Outflow} & 
    \multicolumn{2}{c}{Taxi Inflow} & 
    \multicolumn{2}{c}{Taxi Outflow} \\    
    \cmidrule(lr){2-3} \cmidrule(lr){4-5} \cmidrule(lr){6-7} \cmidrule(lr){8-9}
    
     & MAE & RMSE & MAE & RMSE & MAE & RMSE & MAE & RMSE \\
    \midrule
    
    AGCRN~\cite{bai2020adaptive} &
    0.99 / 1.10 / 1.24 & 1.49 / 1.63 / 1.79 & 1.00 / 1.11 / 1.25 & 1.51 / 1.65 / 1.81 & 2.76 / 3.00 / 3.29 & 4.69 / 5.22 / 5.68 & 2.95 / 3.29 / 3.61 & 5.73 / 6.65 / 7.32 \\
    \textbf{+ FreMo} &
    \textbf{0.95 / 1.07 / 1.20} & \textbf{1.43 / 1.58 / 1.63} & \textbf{0.94 / 0.99 / 1.21} & \textbf{1.42 / 1.46 / 1.76} & \textbf{2.62 / 2.91 / 3.14} & \textbf{4.17 / 4.79 / 5.45} & \textbf{2.80 / 3.08 / 3.26} & \textbf{5.43 / 6.25 / 7.13} \\
    \midrule
  
    TimesNet~\cite{wu2023timesnet} &
    0.54 / 0.59 / 0.62 & 1.17 / 1.27 / 1.32 & 0.56 / 0.60 / 0.63 & 1.20 / 1.30 / 1.35 & 2.55 / 2.78 / 2.94 & 4.38 / 4.87 / 5.58 & 2.79 / 3.04 / 3.22 & 5.41 / \textbf{5.92} / 7.26 \\
    \textbf{+ FreMo} &
    \textbf{0.52 / 0.58 / 0.61} & \textbf{1.15 / 1.23 / 1.28} & \textbf{0.55 / 0.58 / 0.60} & \textbf{1.17 / 1.26 / 1.32} & \textbf{2.50 / 2.74 / 2.84} & \textbf{4.34 / 4.80 / 5.06} & \textbf{2.63 / 2.89 / 3.02} & \textbf{5.10 / 5.63 / 5.89} \\
    \midrule
    
    iTransformer~\cite{liu2024itransformer} &
    0.63 / 0.71 / 0.77 & 1.38 / 1.55 / 1.65 & 0.64 / 0.71 / 0.77 & 1.41 / 1.55 / 1.66 & 3.10 / 3.70 / 4.24 & 5.35 / 6.60 / 7.58 & 3.26 / 4.03 / 4.60 & 6.37 / 8.08 / 9.33 \\
    \textbf{+ FreMo} &
    \textbf{0.62 / 0.67 / 0.72} & \textbf{1.37 / 1.50 / 1.59} & \textbf{0.63 / 0.68 / 0.72} & \textbf{1.35 / 1.44 / 1.60} & \textbf{3.08 / 3.61 / 4.10} & \textbf{5.30 / 6.49 / 7.36} & \textbf{3.21 / 3.79 / 4.01} & \textbf{6.31 / 7.88 / 9.12} \\
    \midrule

    STAEformer~\cite{liu2023spatio} & 0.64 / 0.71 / 0.75 & 1.23 / 1.31 / 1.35 & 0.65 / 0.71 / 0.78 & 1.28 / 1.36 / 1.38 & 2.94 / 3.31 / 3.70 & 5.36 / 6.23 / 6.95 & 3.36 / 3.52 / 3.81 & 7.71 / 7.94 / 8.51 \\
    
    \textbf{+ FreMo} & \textbf{0.62} / \textbf{0.67} / \textbf{0.73} & \textbf{1.19} / \textbf{1.29} / \textbf{1.31} & \textbf{0.62} / \textbf{0.68} / \textbf{0.75} & \textbf{1.26} / \textbf{1.34} / \textbf{1.35} & \textbf{2.75} / \textbf{3.13} / \textbf{3.38} & \textbf{5.12} / \textbf{5.97} / \textbf{6.56} & \textbf{2.96} / \textbf{3.33} / \textbf{3.57} & \textbf{6.76} / \textbf{7.67} / \textbf{8.23} \\
    \midrule
    \midrule

    \multirow{2}{*}{\textbf{Chicago}} & 
    \multicolumn{2}{c}{Bike Inflow} & 
    \multicolumn{2}{c}{Bike Outflow} & 
    \multicolumn{2}{c}{Taxi Inflow} & 
    \multicolumn{2}{c}{Taxi Outflow} \\    
    \cmidrule(lr){2-3} \cmidrule(lr){4-5} \cmidrule(lr){6-7} \cmidrule(lr){8-9}
    
     & MAE & RMSE & MAE & RMSE & MAE & RMSE & MAE & RMSE \\
    \midrule
  
    AGCRN~\cite{bai2020adaptive} & 
    0.51 / 0.52 / 0.60 & 1.35 / 1.52 / 1.75 & 0.52 / 0.53 / 0.60 & 1.45 / 1.57 / 1.73 & 0.76 / 0.79 / 0.90 & 2.51 / 2.74 / 3.18 & 0.69 / 0.73 / 0.85 & 2.53 / 2.87 / 3.44 \\
    \textbf{+ FreMo} &
    \textbf{0.40 / 0.42 / 0.45} & \textbf{1.31 / 1.41 / 1.56} & \textbf{0.39 / 0.42 / 0.44} & \textbf{1.35 / 1.46 / 1.56} & \textbf{0.64 / 0.68 / 0.75} & \textbf{2.35 / 2.55 / 2.88} & \textbf{0.58 / 0.62 / 0.70} & \textbf{2.41 / 2.70 / 3.15} \\
    \midrule
    
    TimesNet~\cite{wu2023timesnet} &
    0.34 / 0.36 / 0.39 & 1.25 / 1.38 / 1.56 & 0.35 / 0.37 / 0.39 & 1.35 / 1.43 / 1.54 & 0.59 / 0.64 / 0.69 & 2.56 / 2.88 / 3.26 & 0.52 / 0.57 / 0.63 & 2.66 / 2.99 / 3.54 \\
    \textbf{+ FreMo} &
    \textbf{0.34 / 0.36 / 0.39} & \textbf{1.23 / 1.36 / 1.51} & \textbf{0.34 / 0.36 / 0.37} & \textbf{1.34 / 1.41 / 1.52} & \textbf{0.57 / 0.62 / 0.66} & \textbf{2.52 / 2.78 / 3.10} & \textbf{0.50 / 0.55 / 0.61} & \textbf{2.61 / 2.87 / 3.40} \\
    \midrule
    
    iTransformer~\cite{liu2024itransformer} &
    0.38 / 0.43 / 0.49 & 1.46 / 1.76 / 2.07 & 0.38 / 0.41 / 0.45 & 1.54 / 1.72 / 1.87 & 0.61 / 0.72 / 0.84 & 2.60 / 3.16 / 3.90 & 0.55 / 0.65 / 0.79 & 2.63 / 3.29 / 4.22 \\
    \textbf{+ FreMo} &
    \textbf{0.37 / 0.41 / 0.47} & \textbf{1.40 / 1.63 / 1.91} & \textbf{0.37 / 0.39 / 0.41} & \textbf{1.44 / 1.58 / 1.75} & \textbf{0.58 / 0.68 / 0.74} & \textbf{2.57 / 2.98 / 3.60} & \textbf{0.54 / 0.60 / 0.71} & \textbf{2.59 / 3.06 / 3.80} \\
    \midrule

    STAEformer~\cite{liu2023spatio} & 0.35 / 0.38 / 0.31 & 1.91 / 1.92 / 1.96 & 0.37 / 0.38 / 0.42 & 1.98 / 2.01 / 2.27 & 0.74 / 0.78 / 0.83 & 4.65 / 4.93 / 5.20 & 0.95 / 0.99 / 1.03 & 4.68 / 4.87 / 5.06 \\
    
    \textbf{+ FreMo} & \textbf{0.34} / \textbf{0.35} / \textbf{0.37} & \textbf{1.68} / \textbf{1.88} / \textbf{1.91} & \textbf{0.35} / \textbf{0.38} / \textbf{0.40} & \textbf{1.77} / \textbf{1.91} / \textbf{2.07} & \textbf{0.60} / \textbf{0.66} / \textbf{0.74} & \textbf{2.99} / \textbf{3.46} / \textbf{4.27} & \textbf{0.54} / \textbf{0.58} / \textbf{0.66} & \textbf{3.07} / \textbf{3.25} / \textbf{3.79} \\
    \bottomrule
    \bottomrule
    \end{tabular}%
}
\label{tab:generality_full}
\end{table*}

\begin{figure*}[t]
  \centering
  \includegraphics[width=0.95\linewidth]{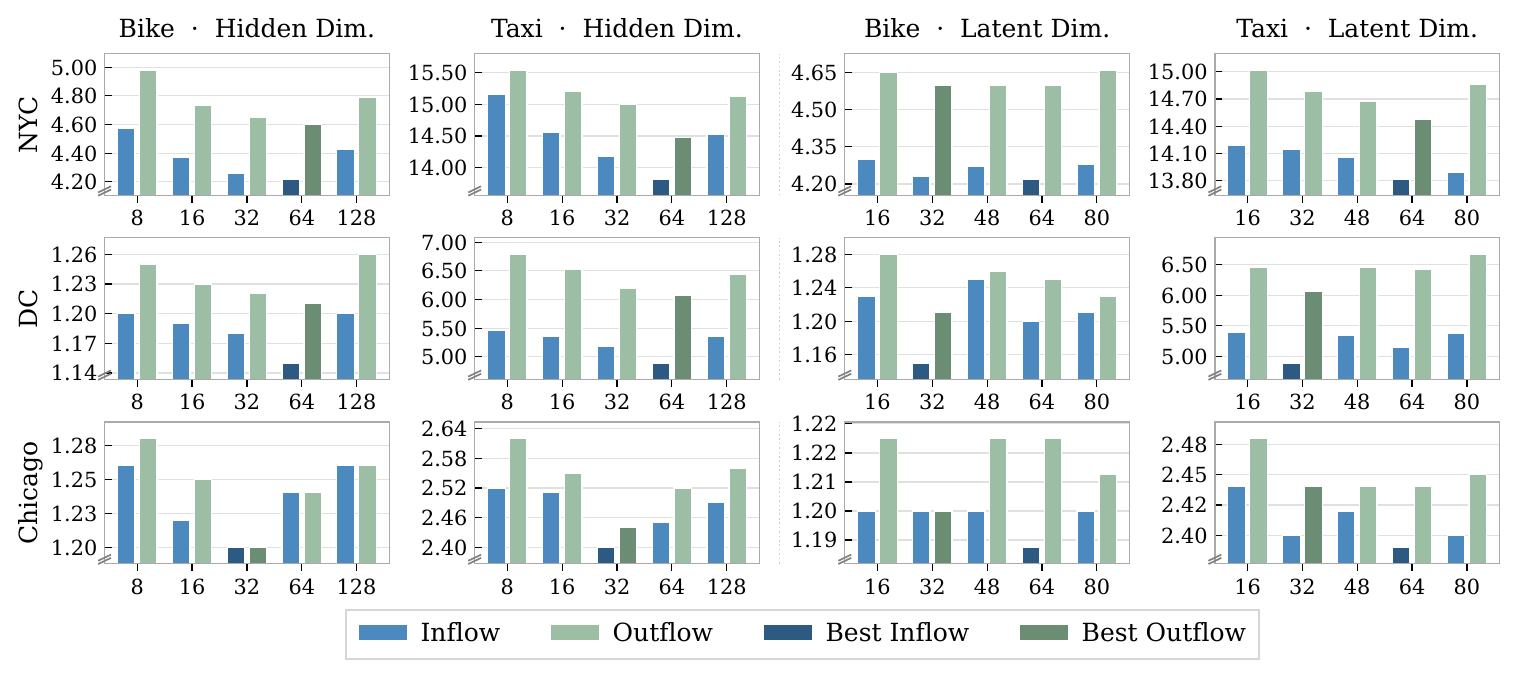}
  \caption{Hyperparameter sensitivity studies of hidden dimension $d$ and latent dimension $L$ on three datasets.}
  \Description{The curves of hyper-parameter studies.}
  \label{figure:hyper_parameter}
\end{figure*}

\section{Full Results of Plug-and-Play Capacity} \label{appendix_generality}
Table~\ref{tab:generality_full} reports the full generality results of integrating FreMo into four representative forecasting backbones (AGCRN, TimesNet, iTransformer, and STAEformer) across all datasets. Consistent with the observations in the main text, FreMo yields universal performance improvements across diverse datasets, backbone architectures, and transportation modalities, indicating FreMo's adaptability to diverse urban mobility patterns and spatial configurations, regardless of the scale or density of the road network. This universality implies that explicit spectral disentanglement and synergy modeling capture fundamental cross-modality correlations that remain under-exploited by current time series modeling paradigms. These extensive results further validate the robustness and effectiveness of FreMo as a general, plug-and-play enhancement module for multi-modality transportation forecasting.

\section{Full Results of Hyperparameter Sensitivity} \label{appendix_hyperparameter}
To investigate the impact of hyperparameter settings on FreMo, we conduct a sensitivity analysis on two key parameters: the hidden dimension ($d$) and the latent dimension ($L$). We evaluate the RMSE metric across three datasets by varying $d$ within $\{8, 16, 32, 64, 128\}$ and $L$ within $\{16, 32, 48, 64, 80\}$. The results are visualized in Figure~\ref{figure:hyper_parameter}.
Regarding the hidden dimension $d$, we observe a consistent trend where performance improves with increased model capacity but degrades at higher values (e.g., 128) due to overfitting. NYC and Washington DC achieve optimal results at $d=64$, maximizing spectral feature encoding. In contrast, the Chicago dataset favors a more compact representation, peaking at $d=32$. 
For the latent dimension $L$, the results indicate that a moderate embedding size is sufficient to capture spatial heterogeneity without introducing redundancy. Performance generally stabilizes around $L=64$ for NYC. However, Washington DC and Chicago exhibit a preference for a lower dimension, achieving its lowest error at $L=32$. Based on these empirical observations, we adopt the optimal $(d, L)$ configurations of $(64, 64)$ for NYC, $(64, 32)$ for Washington DC, and $(32, 32)$ for Chicago in our main experiments (Table~\ref{tab:performance}).

\end{document}